\documentclass[lettersize,journal]{IEEEtran}
\usepackage{amsmath,amsfonts,amssymb,amsthm}
\usepackage{array}
\usepackage[caption=false,font=normalsize,labelfont=sf,textfont=sf]{subfig}
\usepackage{textcomp}
\usepackage{stfloats}
\usepackage{url}
\usepackage{verbatim}
\usepackage{bm}
\usepackage{graphicx}
\usepackage{multirow}
\usepackage{cite}
\usepackage[utf8]{inputenc}
\usepackage{mathrsfs}
\usepackage[linesnumbered,boxed,ruled,commentsnumbered]{algorithm2e}
\usepackage{algorithmic}
\usepackage{xcolor}
\usepackage{float}

\usepackage[hidelinks]{hyperref}
\DeclareMathOperator{\clip}{clip}
\newcommand{\orcid}[1]{\href{https://orcid.org/#1}{\includegraphics[height = 2ex]{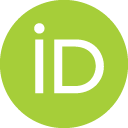}}}
\newcolumntype{C}[1]{>{\centering\arraybackslash}p{#1}}
\newcolumntype{M}[1]{>{\centering\arraybackslash}m{#1}}
\begin{document}

\title{Boosting Adversarial Attacks by Leveraging Decision Boundary Information}

\author{Boheng Zeng\orcid{0000-0003-3248-5324}, LianLi Gao, QiLong Zhang, ChaoQun Li, JingKuan Song\orcid{0000-0002-2549-8322} and ShuaiQi Jing
\thanks{Boheng Zeng, LianLi Gao, QiLong Zhang, ChaoQun Li, JingKuan Song and ShuaiQi Jing are with University of Electronic Science and Technology of China, ChengDu 611730, China (e-mail: boheng.zeng@std.uestc.edu.cn; lianli.gao@uestc.edu.cn; qilong.zhang@std.uestc.edu.cn; lichaoqunuestc@gmail.com; jingkuan.song@gmail.com; jingshuaiqi@uestc.edu.cn)}
}



\maketitle

\begin{abstract}

Due to the gap between a substitute model and a victim model, the gradient-based noise generated from a substitute model may have low transferability for a victim model since their gradients are different. Inspired by the fact that the decision boundaries of different models do not differ much, we conduct experiments and discover that the gradients of different models are more similar on the decision boundary than in the original position. 
Moreover, since the decision boundary in the vicinity of an input image is flat along most directions, we conjecture that the boundary gradients can help find an effective direction to cross the decision boundary of the victim models. 
Based on it, we propose a Boundary Fitting Attack to improve transferability. 
Specifically, we introduce a method to obtain a set of boundary points and leverage the gradient information of these points to update the adversarial examples. 
Notably, our method can be combined with existing gradient-based methods. Extensive experiments prove the effectiveness of our method, \textit{i.e.}, improving the success rate by \textbf{5.6\%} against normally trained CNNs and \textbf{14.9\%} against defense CNNs on average compared to state-of-the-art transfer-based attacks. 
Further we compare transformers with CNNs, the results indicate that transformers are more robust than CNNs.
However, our method still outperforms existing methods when attacking transformers.
Specifically, when using CNNs as substitute models, our method obtains an average attack success rate of 58.2\%, which is 10.8\% higher than other state-of-the-art transfer-based attacks. 
\end{abstract}


\section{Introduction}

Even though convolutional neural networks (CNNs) have achieved significant success, Szegedy \textit{et al.}~\cite{Intriguing} find that adding human-imperceptible noises to clean images can easily fool state-of-the-art CNNs. Worse still, the malicious adversarial examples can be applied to cause social harm in real-world applications like face recognition~\cite{deepface,tcsvtfacerecognition,ArcFace} and self-driving cars~\cite{SocialGAN}, which pose a severe threat to CNNs. To further explore the vulnerability of CNNs, various works~\cite{fgsm,ifgsm,gao2021feature,tcsvtcameraattack,tcsvttargetattack} pay attention to the generation of adversarial examples.

\begin{figure}[t]
    \centering
    \includegraphics[width=8cm]{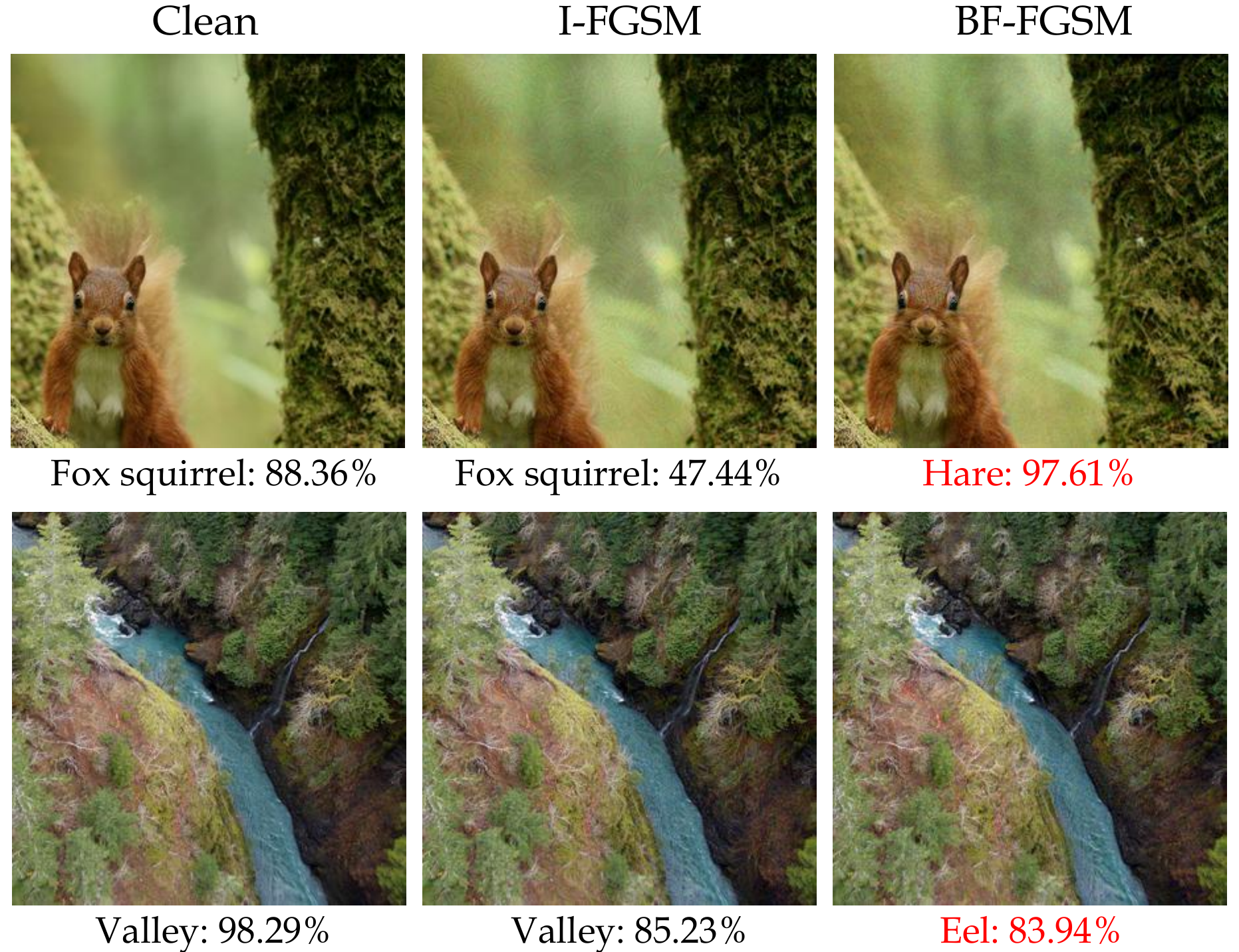}
    \caption{Adversarial examples crafted via Inc-v3 by our BF-FGSM transfer towards victim model Res-152. The bottom caption of each image is its predicted label and corresponding confidence.}
    \label{fig:vis_compare}
\end{figure}

\begin{figure*}[t]
    \centering
    \includegraphics[height=7cm]{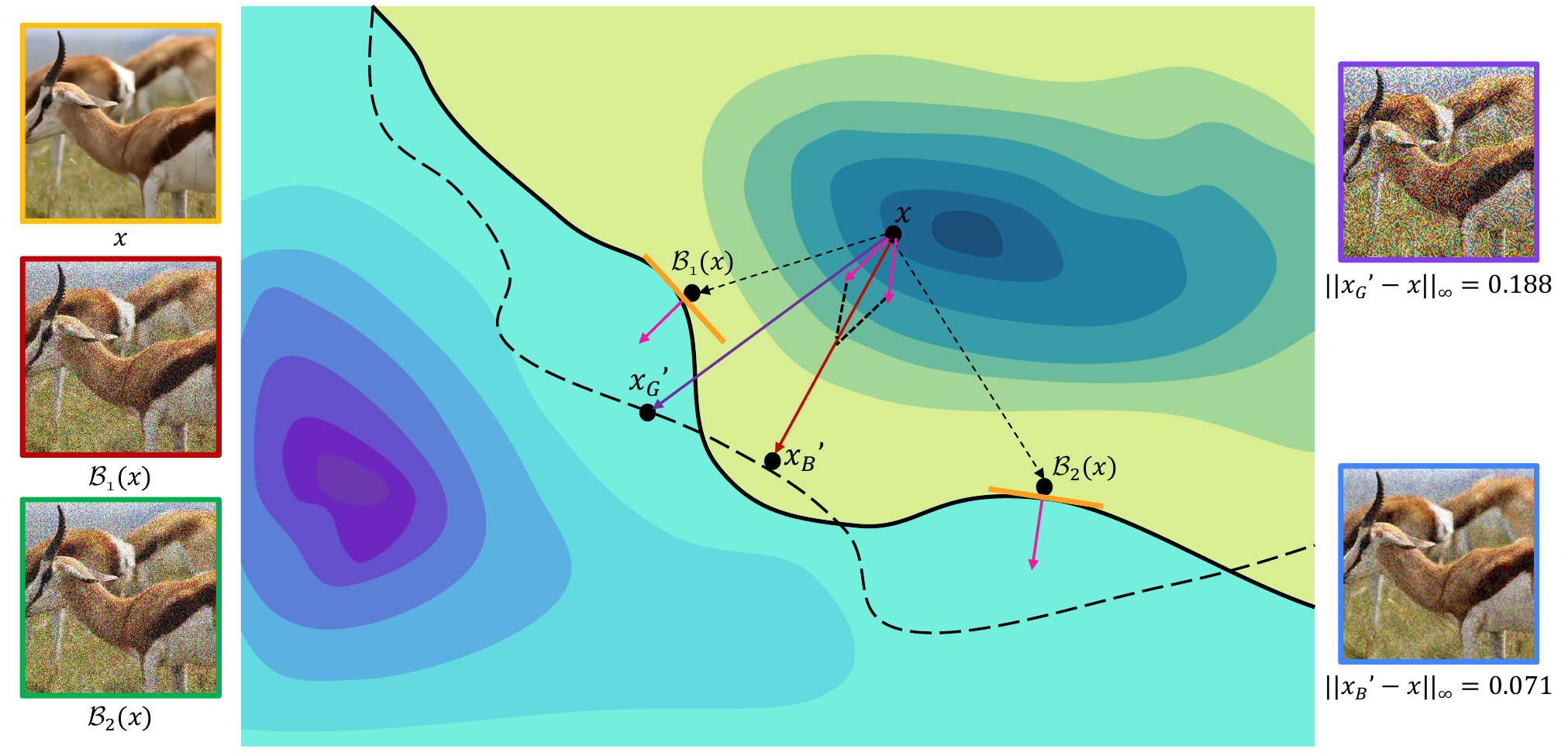}
    \caption{Overview of our Boundary Fitting Method. Background is contour of cross entropy loss. 
    The solid black line is the decision boundary, which separate 'gazelle' and other categories, for the substitute model Inc-v3 while the dashed black line is of the victim model Inc-v4. 
    We sample a batch of boundary points, \textit{i.e.} $\mathcal{B}_1(x)$ and $\mathcal{B}_2(x)$ (in fact, we sampled 20 boundary points in our experiment as default), and average their gradients.
    The generated direction by our method (red line) is easier to fool Inc-v4 than the direction of the original gradient of $\bm{x}$ (purple line), as the left-hand plot shows that our generated direction has a shorter distance to the decision boundary. }
    \label{fig:overview}
\end{figure*}

Generally, adversarial attack settings can be broadly divided into white-box and black-box. In the white-box setting~\cite{cw,deepfool,curl&whey}, an attacker has full knowledge of model architecture and parameters, thus having a high attack success rate. However, white-box attacks are impracticable for real-world applications since it is almost impossible for an attacker to obtain such information from the victim model. To tackle this issue, most of the black-box methods resort to the cross-model transferability~\cite{transferability} to perform attacks, \textit{i.e.}, adversarial example crafted by a substitute model 
may fool other models.

Nevertheless, the gap (\textit{e.g.}, loss surface) between models is usually large, which limits the transferability of adversarial examples.
To narrow the gap, various methods~\cite{mifgsm,difgsm,pifgsm,sinifgsm,fia,admix} are proposed recently. 
For instance, Gao \textit{et al.}~\cite{pifgsm} propose a patch-wise iterative algorithm to update adversarial examples.
Xie \textit{et al.}~\cite{difgsm} apply random sizing and padding to transform images. 
Wang \textit{et al.}~\cite{admix} propose a method that admixes the input image with a set of images randomly sampled from other categories as input, where the mixed images retain the original label. 
However, these methods may not find effective data points for calculating gradients, thus limiting the transferability of adversarial examples.

However, CNNs share many commonalities, e.g decision boundary, which enabling transferability.
Tram{\`{e}}r \textit{et al.} ~\cite{AdversarialSubspace} find that adversarial subspace span a contiguous subspace of large dimensionality and different models share a significant fraction of the adversarial subspace thus enabling transferability.
Therefore, finding a highly overlapped adversarial subspace can boost attacks.
Moreover, based on the observation~\cite{AdversarialSubspace} of the decision boundaries of the different models are actually close, we further demonstrate that different models have more similar gradients on the decision boundary than on the original position.
Besides, as the prior work ~\cite{EmpiricalGeometry} shows that the decision boundary tends to be flat,
we conjecture that leveraging the decision boundary gradient of the substitute mode can find an effective direction across the decision boundary of different models.

Based on the analysis above, we propose a Boundary Fitting Attack to enhance the transferability of adversarial examples. 
The details are shown in Algorithm.~\ref{alg:Framwork}. 
In conjunction with Figure~\ref{fig:overview}, we introduce our method. 
First, we move the input image $\bm{x}$ along random directions to obtain a set of boundary points. 
Then we average the gradients of these boundary points to get the update direction, \textit{i.e.}, the red line. 
Compared with the purple line, which denotes the original gradient direction of $\bm{x}$, our derived direction crosses the decision boundary of the victim model at a closer distance (experimental results shown in Table~\ref{tab:l2explantation} also support it).
Further, we demonstrate theoretically and experimentally the effectiveness of boundary gradient averaging and give information about the time consumption of this operation.

To reflect the boundary distance on transferability, we compare the proposed Boundary Fitting Attack with state-of-the-art attack methods on the ImageNet dataset. Notably, our attack achieves the highest average attack success rate against both normally trained and defense CNNs. In addition, our attack also has best performance on transformers whatever from CNNs transfer to transformers (CNNs2trans), transformers to transformers (trans2trans) or transformers to CNNs (trans2CNNs). And the experiments show that the average attack success rate of CNNs2trans lower than trans2CNNs which manifest transformers are more robust than CNNs in a some certain.
To sum up, our contributions are as follows:


\begin{itemize}
    \item We analyze the decision boundaries of different models and find that the gradients at the boundary points are more similar than the original gradients between models. These gradients on the decision boundary can potentially form a transferable perturbation noise. Based on the analysis, we propose a boundary fitting attack, which averages the gradients of a set of boundary points to generate a direction that can effectively cross the decision boundaries of victim models. 
    \item We propose a concept: decision boundary distance. We conduct experiments and find that robustness is positively correlated with the decision boundary distance along natural directions, which also explains why transformers are more robust than CNNs.
    \item Extensive experiments on the ImageNet dataset show that our proposed method significantly improves the success rates by 5.6\% and 14.9\% on normally trained and defense CNN models, respectively. We also compared transformers with CNNs, the results indicate that transformers are more robust than CNNs. Moreover, our method have even higher performance on transformers. Our method outperforms other state-of-the-art transfer methods by 10.8\%, 11.1\% and 13.6\% on CNNs2trans, trans2trans and trans2CNNs respectively.
\end{itemize}

\section{Related Work}

\subsection{Adversarial Attacks}

Since Goodfellow \textit{et al.}~\cite{fgsm} proposed Fast Gradient Sign Method (FGSM) to craft adversarial examples, various works proposed novel methods based on it to improve transferability.
Therefore, we follow this family to boost adversarial attacks.
Dong \textit{et al.}~\cite{mifgsm} introduce the momentum term to stabilize update directions and avoid adversarial examples falling into local optimum.
Xie \textit{et al.}~\cite{difgsm} propose to improve the transferability of adversarial examples by applying diverse input transformation. 
Dong \textit{et al.}~\cite{tifgsm} shift the input to create diverse translated images and approximately estimate the overall gradient to mitigate the problem of over-reliance on the substitute model. 
Gao \textit{et al.}~\cite{pifgsm} propose patch-wise perturbations to better cover the discriminate region of images.
Lin \textit{et al.}~\cite{sinifgsm} discover the scale-invariant property of DNNs and thus average the gradients of different scaled images to update adversarial examples.
Wang \textit{et al.}~\cite{vtfgsm} stabilize the update direction by fine-tuning the current gradient based on the gradient variance from the last iteration. 
Wang \textit{et al.}~\cite{fia} disrupt important features which dominate model predictions and promote trivial features to boost attacks.
Wang \textit{et al.}~\cite{admix} average the gradients of a set of admixed images (\textit{i.e.}, an input image admixed with a small portion of other images) to boost attack.

\subsection{Decision Boundary of DNN models}

Recently, several works have analyzed the decision boundary of DNNs. 
Fawzi \textit{et al.} ~\cite{RobustnessOfClassifiers} analyze the robustness of DNNs from a perspective of the curvature of decision boundary, and suggest that DNNs are robust to random noise when the decision boundary has a small curvature. 
Tram{\`{e}}r \textit{et al.} ~\cite{AdversarialSubspace} propose novel methods for estimating the space of adversarial examples. They find that the decision boundary between different models are actually close in arbitrary directions, and the adversarial subspace share a large fraction between different models which enabling transferability.
Fawzi \textit{et al.}~\cite{EmpiricalGeometry} experimentally found that the classification region is connected and the decision boundary is flat in most directions, which makes it possible to obtain boundary points. 
Khoury \textit{et al.}~\cite{GeometryAE} highlight the importance of codimension: for an input image, it can be represented as low-dimension data manifold in high-dimension space and there are many directions off the manifold construct adversarial examples. Furthermore, they propose a gradient-free geometric attack to manifest the importance of the decision boundaries. 
Maho \textit{et al.} ~\cite{SurFree} proposed a query-based attack SurFree which focuses on the geometrical properties of the decision boundary and significantly reduce the query budgets. 
While recent works almost focus on the decision boundary of a single model. 
The decision boundary relationships between models have rarely been studied. 
In this paper, we focus on using the decision boundary of the substitute model to fit other models thus improve the transferability of adversarial examples.

\subsection{Adversarial Defenses} 

As a counterpart of adversarial attacks, adversarial defenses aim to mitigate the threat of adversarial examples. 
To that end, various adversarial defense methods have been proposed in recent years. 
Tram{\`{e}}r \textit{et al.}~\cite{ens} introduce an ensemble adversarial training method that augments training data with adversarial examples crafted by other models. 
Xie \textit{et al.}~\cite{RP} apply random resizing and padding (RP) to the inputs at inference time to mitigate adversarial effects. 
Liao \textit{et al.}~\cite{HGD} propose high-level representation guided denoiser (HGD) to suppress the influence of adversarial perturbation. 
Xie \textit{et al.}~\cite{FeatureDeoise} develop an end-to-end trained network with the aim of denoising the intermediate features to significantly improve the robustness against adversarial exampels.
Jia \textit{et al.}~\cite{RS} leverage randomized smoothing (RS) with Gaussian noise to enhance model robustness.
Naseer \textit{et al.}~\cite{NRP} design a Neural Representation Purifier (NRP) model that learns to clean adversarial perturbed images based on a self-supervised adversarial training mechanism.
Wang \textit{et al.}~\cite{tcsvtdefense} build a more robust classification system that can be viewed as a structural black box. After adding a buffer to the classification system, attackers can be efficiently deceived.

\section{Methodology}
In this section, we give an introduction to our motivation and the implementation of our approach. We first give a brief definition of adversarial attacks and the decision boundary in Sec.~\ref{sec:3.1}. In sec.~\ref{sec:3.2}, we introduce our motivation and propose a conjecture. Based on it, we finally provide the detailed algorithm of our method in Sec.~\ref{sec:3.3}.

\subsection{Preliminaries}
\label{sec:3.1}

Formally, let $\bm{x}$ denote an image, and $y$ denote the corresponding ground-truth label. We use $f_\theta: \bm{x} \rightarrow y$ denote a classification model, where $\theta$ indicates the parameters of the model. 
The goal of the adversarial attacks is to seek an adversarial example $\bm{x'}$ within a $l_p$-ball around the original image to mislead the classifier, \textit{i.e.}, $f_\theta(\bm{x'})\neq y$ ($a.k.a.$ non-targeted attack). 
Following prior works~\cite{fgsm,mifgsm,fia}, we use $l_\infty$-norm to constrain the size of perturbation, \textit{i.e.}, $||\bm{x'}-\bm{x}||_\infty\leq \epsilon$ and $\epsilon$ is the maximum perturbation. 

For a classifier which has a $k$ classification regions $\mathcal{R}$, \textit{e.g.}, $\mathcal{R}_i$ corresponds to the classification region of class $i$. The decision boundary $\mathcal{B}_{i,j}$ separates $\mathcal{R}_i$ and $\mathcal{R}_{j}$, which can be expressed as follows:
\begin{equation}
    \mathcal{B}_{i,j} = \{f_i(\bm{x}) - f_{j}(\bm{x}) = 0 \}.
\end{equation}

In this paper, we focus on non-targeted attacks. Therefore, we are only concerned with the decision boundary $\mathscr{B}_{i}$ which can be expressed as
\begin{equation}
    \mathcal{B}_{i} = \mathop{\cup}\limits_{j=1}^k \mathcal{B}_{i,j}, j\neq i,
\end{equation}

\subsection{Motivation}
\label{sec:3.2}


Motivated by prior work~\cite{AdversarialSubspace} which shows that the decision boundaries of different models are actually close, we wonder whether the gradients on the decision boundary points of different models are more similar than the gradients on the original position.
To check it, we evaluate the gradient cosine similarity of several white-box$\rightarrow$black-box pairs at different data points. 

\begin{table}[h]
\centering
    \caption{The average of cosine similarity of the gradients between the substitute model and the victim model. For the boundary gradient of the substitute model, we use Eq.~\ref{eq:movetoboundary} and Eq.~\ref{eq:updatedirection} to approximate the decision boundary and calculate the (average) gradient $\mathcal{G}$ on resulting boundary point(s). Along this direction $\mathcal{G}$, we proceeded to estimate decision boundary of the victim model and obtain the gradient.}
    \label{tab:motivation}
    \resizebox{1\linewidth}{!}{
    \begin{tabular}{ccc}
        \hline
         Gradients & Inc-v3 $\rightarrow$ Inc-v4 & Inc-v3 $\rightarrow$ IncRes-v2$_{ens}$\\
         \hline
         Original gradient& 1.6e$^{-2}$ & 2.1e$^{-3}$ \\
         Boundary gradient (N=1)& 1.8e$^{-2}$ (12.5\% $\uparrow$) & 2.6e$^{-3}$  (23.8\% $\uparrow$)\\
         Boundary gradient (N=20)& 2.3e$^{-2}$ (43.8\% $\uparrow$) & 3.0e$^{-3}$  (42.9\% $\uparrow$)\\
         \hline
    \end{tabular}
    }
\end{table}


As demonstrated in Table.~\ref{tab:motivation}, boundary gradients are less diverse, and there is more commonality between the different models.
Inspired by the above observation and the discovery~\cite{EmpiricalGeometry} that the decision boundary tends to be flat, we have the following conjecture: \textit{Leveraging the decision boundary gradient of the substitute model can generate a more effective direction to cross the decision boundaries of victim models.}

\subsection{Boundary Fitting Attack}
\label{sec:3.3}

Based on the above conjecture, we propose a Boundary Fitting Attack that leverages the boundary gradients of the substitute model to enhance the transferability of resulting adversarial examples, as detailed in the Algorithm.~\ref{alg:Framwork}.  Specifically, we first introduce a random transformation $\mathcal{B}_i(\cdot)$ which moves an image $\bm{x}: f(\bm{x})=y$ to its decision boundary $\mathcal{B}_{f(x)}$:
\begin{equation}
    \mathcal{B}_i(\bm{x})=\bm{x}+\gamma^t \cdot \bm{d_i},\,\,\,\,\text{ s.t.  $f(\mathcal{B}_i(\bm{x}))=f(\bm{x})$}, 
\label{eq:movetoboundary}
\end{equation}
here we randomly choose a direction $d_i$ which is sampled from a Gaussian distribution $\mathcal{N} (0, \sigma^{2} I)$ to keep diversity. Here we set $\sigma$ to a large value so that $\bm{x}+\bm{d_i}$ is capable of moving out of $\mathcal{B}_{f(x)}$. After that, we repeatedly multiply a shrinkage factor $\gamma$ until it goes back to the source classification region.

Since Table~\ref{tab:motivation} demonstrates that averaging the gradients of multiple boundary points can better fit the black-box model, we obtain the update direction by a set of boundary points \{ $\mathcal{B}_1, ..., \mathcal{B}_n$ \} (\textit{i.e.}, apply $\mathcal{B}_i(\cdot)$ multiple times) with the aim of enhancing the transferability of adversarial examples. Formally, it can be expressed as:
\begin{equation}
    \mathcal{G} = \frac{1}{N} \sum_{i=1}^N \nabla_{\mathcal{B}_i(\bm{x})}J(\mathcal{B}_i(\bm{x}),y;\phi).
\label{eq:updatedirection}
\end{equation}

In combination with I-FGSM, the update function of our proposed Boundary Fitting Fast Gradient Sign Method (BF-FGSM) can be written as:
\begin{equation}
    \bm{x'}_{t+1} = \clip_{\bm{x}, \epsilon}\{\bm{x'}_t + \alpha \cdot sign(\mathcal{G}))\},
\end{equation}
\noindent where $\clip_{\bm{x}, \epsilon}(\cdot)$ denotes an element-wise clipping operation to ensure $\bm{x'}\in[\bm{x}-\epsilon,  \bm{x}+\epsilon]$, and $\alpha$ is the step size. The adversarial examples are shown in Figure~\ref{fig:vis_compare}. 
Compared with I-FGSM, our method yields more transferable adversarial examples, which can fool black-box models with high confidence.

\begin{algorithm}[t]
\caption{BF-FGSM}
\label{alg:Framwork} 
\SetAlgoNoLine
\SetKwInOut{Input}{\textbf{Input}}\SetKwInOut{Output}{\textbf{Output}}
\LinesNumbered
\Input{A classifier $f$ with parameters $\phi$; loss function $J$; a clean image \bm{$x$} with ground-truth label $y$; iterations $T$; $L_\infty$ constraint $\epsilon$; integrated boundary points $N$; shrinkage factor $\gamma$.}
\Output{The adversarial example \bm{$x'$}}
$\alpha = \epsilon / T$, $\bm{x'_0} = \bm{x}$\\
\For{$t=0 \rightarrow T-1$}{
\For{$i = 1 \rightarrow N$}{
Get boundary point $\mathcal{B}_i(\bm{x'_{t}})$
using Eq.~\ref{eq:movetoboundary}\\
Gradient integration $\mathcal{G}_i=\nabla_{\mathcal{B}_i(\bm{x'_{t}})} J(\mathcal{B}_i(\bm{x'_{t}}), y; \phi)$ 
}
Average gradient: $\mathcal{G} = \frac{1}{N}\sum_{i=1}^{N}{\mathcal{G}_i}$\\
$\bm{x'_{t+1}} = \clip_{x,\epsilon}\left\{\bm{x'_{t}} + \alpha \cdot sign(\mathcal{G})\right\}$\\
$\bm{x'_{t+1}} = \clip(\bm{x'_{t+1}}, 0, 1)$\\
}
$\bm{x'}=\bm{x'_T}$\\
\Return $\bm{x'}$
\end{algorithm}

\begin{table}[h]
\centering
\caption{The average $\mathcal{L}_\infty$-distance between boundary points and input images on each victim model (\textit{i.e.}, Inc-v4, IncRes-v2, Res-152 and IncResv2$_{ens}$.), where the boundary points are obtained by moving along the direction of the sign gradient generated via the substitute model Inc-v3 by I-FGSM, DI-FGSM, TI-FGSM, SI-FGSM, FI-FGSM, Admix and BF-FGSM at the first step, respectively.}
\resizebox{0.98\linewidth}{!}{
\begin{tabular}{ccccc}
\hline
Attacks & Inc-v4 & IncRes-v2 & Res-152 & IncResv2$_{ens}$ \\
\hline
I-FGSM & 39.64 & 40.15 & 35.70 & 54.73 \\
DI-FGSM & 38.85 & 39.11 & 35.75 & 53.92 \\
TI-FGSM & 32.36 & 33.86 & 30.78 & 43.37    \\
SI-FGSM & 27.66 & 26.90 & 25.88 & 48.14    \\
FI-FGSM & 24.73 & 24.86 & 26.16 & 48.94 \\
Admix & 25.54 & 25.56 & 25.17 & 47.27   \\
BF-FGSM (Ours) & \textbf{21.17} & \textbf{19.23} & \textbf{21.81} & \textbf{41.60}    \\
\hline
\end{tabular}}

\label{tab:l2explantation}
\end{table}

\begin{table*}[t]
\centering
\caption{The attack success rates (\%) of black-box attacks against six normally trained models. The adversarial examples are crafted via Inc-v3, Inc-v4, IncRes-v2 and Res-152, respectively. "*" indicates white-box attacks.}
\resizebox{1\linewidth}{!}{
\begin{tabular}{c|c|M{1cm}|M{1cm}|M{1cm}|M{1cm}|M{1cm}|M{1cm}|M{1cm}}
\hline
Models & Attacks & Inc-v3 & Inc-v4 & IncRes-v2 & Res-50 & Res-101 & Res-152 & Average \\
\hline
\hline
\multirow{8}{*}{Inc-v3} & MI-FGSM & \textbf{100*} & 50.6 & 47.2 & 46.9 & 41.7 & 40.6 & 54.5 \\
 & DI-FGSM & 99.7* & 48.3 & 38.2 & 39.0 & 33.8 & 31.8 & 48.5 \\
 & PI-FGSM & 100* & 55.7 & 49.3 & 51.8 & 47.3 & 45.6 & 58.3 \\
 & BF-FGSM (Ours) & 98.9* & \textbf{64.2} & \textbf{59.1} & \textbf{56.4} & \textbf{51.1} & \textbf{48.4} & \textbf{63.0} \\
\cline{2-9}
 & SI-MI-FGSM & \textbf{100*} & 69.6 & 68.6 & 67.9 & 64.0 & 62.2 & 72.1 \\
 & VT-MI-FGSM & \textbf{100*} & 74.9 & 70.6 & 67.2 & 63.8 & 63.6 & 73.4 \\
 & FI-MI-FGSM & 96.4* & 84.7 & 80.1 & 78.0 & 75.4 & 71.8 & 81.1 \\
 & BF-MI-FGSM (Ours) & 99.1* & \textbf{88.9} & \textbf{88.8} & \textbf{84.5} & \textbf{83.9} & \textbf{85.1} & \textbf{88.4} \\
 \hline
\multirow{8}{*}{Inc-v4} & MI-FGSM & 62.0 & \textbf{100*} & 46.2 & 47.7 & 42.8 & 41.4 & 56.7 \\
 & DI-FGSM & 54.1 & 99.1* & 36.3 & 33.7 & 30.4 & 31.4 & 47.5 \\
 & PI-FGSM & 55.5 & 96.3* & 38.5 & 44.5 & 38.0 & 39.0 & 52.0 \\
 & BF-FGSM (Ours) & \textbf{73.3} & 98.2* & \textbf{60.3} & \textbf{55.3} & \textbf{50.5} & \textbf{49.8} & \textbf{64.6} \\
 \cline{2-9}
 & SI-MI-FGSM & 83.6 & \textbf{100*} & 75.0 & 72.7 & 68.9 & 69.8 & 78.3 \\
 & VT-MI-FGSM & 79.6 & 99.9* & 70.8 & 64.5 & 63.8 & 63.6 & 73.7 \\
 & FI-MI-FGSM & 83.9 & 94.9* & 78.8 & 76.3 & 75.9 & 73.3 & 80.5 \\
 & BF-MI-FGSM (Ours) & \textbf{90.7} & 98.3* & \textbf{86.8} & \textbf{83.8} & \textbf{82.6} & \textbf{83.1} & \textbf{87.6} \\
 \hline
\multirow{8}{*}{IncRes-v2} & MI-FGSM & 60.4 & 52.8 & 99.4* & 49.1 & 46.3 & 45.9 & 59.0 \\
 & DI-FGSM & 56.5 & 49.1 & 97.8* & 38.3 & 37.1 & 35.6 & 52.4 \\
 & PI-FGSM & 64.8 & 58.7 & 99.7* & 53.3 & 48.5 & 48.4 & 62.2 \\
 & BF-FGSM (Ours) & \textbf{74.9} & \textbf{68.5} & 95.2* & \textbf{61.2} & \textbf{58.3} & \textbf{56.6} & \textbf{69.1} \\
 \cline{2-9}
 & SI-MI-FGSM & 84.5 & 83.1 & \textbf{99.7*} & 79.2 & 78.7 & 76.4 & 83.6 \\
 & VT-MI-FGSM & 80.8 & 77.6 & 99.3* & 69.8 & 69.2 & 66.4 & 77.2 \\
 & FI-MI-FGSM & 81.7 & 77.5 & 88.9* & 74.7 & 75.1 & 71.8 & 78.3 \\
 & BF-MI-FGSM (Ours) & \textbf{89.3} & \textbf{87.7} & 95.6* & \textbf{84.4} & \textbf{83.4} & \textbf{83.6} & \textbf{87.3} \\
 \hline
\multirow{8}{*}{Res-152} & MI-FGSM & 54.7 & 50.1 & 45.5 & 84.0 & 86.5 & 99.4* & 70.0 \\
 & DI-FGSM & 57.3 & 51.5 & 47.2 & 83.1 & 85.1 & 99.3* & 70.6 \\
 & PI-FGSM & 64.2 & 55.1 & 49.0 & 81.4 & 83.9 & 99.6* & 72.2 \\
 & BF-FGSM (Ours) & \textbf{65.0} & \textbf{56.3} & \textbf{56.9} & \textbf{89.9} & \textbf{91.3} & \textbf{99.8*} & \textbf{76.5} \\
 \cline{2-9}
 & SI-MI-FGSM & 73.0 & 69.9 & 66.6 & 93.1 & 94.4 & \textbf{99.8*} & 82.8 \\
 & VT-MI-FGSM & 73.7 & 68.8 & 66.4 & 93.3 & 93.8 & 99.5* & 82.6 \\
 & FI-MI-FGSM & 84.1 & 82.8 & 79.0 & 94.5 & 93.9 & 99.2* & 88.9 \\
 & BF-MI-FGSM (Ours) & \textbf{89.0} & \textbf{86.9} & \textbf{87.5} & \textbf{96.7} & \textbf{97.1} & \textbf{99.8*} & \textbf{92.8}\\
  \hline
\end{tabular}
}
\label{tab:normal}
\end{table*}

\begin{table*}[h]
\centering
\caption{The attack success rates (\%) of black-box attacks against eleven defenses. The adversarial examples are crafted via Inc-v3. Note that Admix is equipped with SI-FGSM by default.}
\resizebox{1\linewidth}{!}{
\begin{tabular}{c|c|M{1cm}|M{1cm}|M{1cm}|M{1cm}|M{1cm}|M{1cm}|M{1cm}|M{1cm}|M{1cm}|M{1cm}|M{1cm}}
\hline
 Models & Attacks & Inc-v3$_{ens3}$ & Inc-v3$_{ens4}$ & IncRes-v2$_{ens}$ & HGD & R\&P & RS & NRP & NIPS-r3 & Res152$_{B}$ & Res152$_{D}$ & ResNeXt$_{DA}$\\
\hline
\hline
\multirow{8}{*}{Inc-v3} & TI-DIM & 44.8 & 43.0 & 29.8 & 36.8 & 30.6 & 56.2 & 22.2 & 38.2 & 2.8 & 2.3 & 1.4 \\
& PI-TI-DIM & 43.7 & 46.9 & 35.6 & 34.1 & 36.2 & 75.2 & 37.4 & 40.2 & 5.8 & 4.2 & 4.6 \\
  & SI-TI-DIM & 67.8 & 64.5 & 50.2 & 56.3 & 51.7 & 65.7 & 40.1 & 61.2 & 6.1 & 4.5 & 4.2 \\
  & VT-TI-DIM & 65.3 & 64.3 & 52.2 & 58.8 & 54.0 & 61.8 & 37.9 & 58.8 & 4.5 & 3.8 & 3.6 \\
    & FI-TI-DIM & 62.3 & 60.4 & 50.0 & 52.1 & 48.6 & 64.8 & 38.4 & 55.6 & 6.3 & 3.6 & 4.3 \\
  & BF-TI-DIM (Ours) & \textbf{83.0} & \textbf{83.6} & \textbf{75.6} & \textbf{78.2} & \textbf{73.1} & \textbf{80.7} & \textbf{69.9} & \textbf{80.0} & \textbf{9.1} & \textbf{8.6} & \textbf{7.9} \\
  \cline{2-13}
  & Admix-TI-DIM & 75.3 & 72.1 & 56.8 & 65.7 & 59.8 & 70.5 & 45.3 & 66.0 & 5.7 & 5.2 & 4.6 \\
& BF-SI-TI-DIM (Ours) & \textbf{89.5} & \textbf{88.1} & \textbf{79.7} & \textbf{81.1} & \textbf{78.8} & \textbf{81.7} & \textbf{78.1} & \textbf{84.3} & \textbf{10.2} & \textbf{9.5} & \textbf{9.3} \\
\hline
\multirow{8}{*}{Inc-v4} & TI-DIM & 38.4 & 38.5 & 27.6 & 34.1 & 29.4 & 55.3 & 19.1 & 33.2 & 2.9 & 1.7 & 1.6 \\
 & PI-TI-DI-FGSM & 42.7 & 44.2 & 32.8 & 33.4 & 34.0 & 75.0 & 32.9 & 37.2 & 5.8 & 4.2 & 6.4 \\
 & SI-TI-DIM & 70.3 & 67.5 & 57.3 & 64.6 & 58.2 & 67.8 & 41.4 & 62.3 & 5.3 & 4.6 & 4.8 \\
 & VT-TI-DIM & 57.9 & 57.5 & 46.7 & 56.1 & 58.0 & 59.4 & 41.6 & 62.4 & 3.5 & 3.0 & 3.3 \\
 & FI-TI-DIM & 61.4 & 58.0 & 51.2 & 54.2 & 51.8 & 62.7 & 39.0 & 55.9 & 6.0 & 3.8 & 3.8 \\
 & BF-TI-DIM (ours) & \textbf{83.7} & \textbf{82.1} & \textbf{74.9} & \textbf{77.3} & \textbf{76.2} & \textbf{79.2} & \textbf{70.9} & \textbf{78.6} & \textbf{9.1} & \textbf{7.4} & \textbf{8.4} \\
 \cline{2-13}
  & Admix-TI-DIM & 77.3 & 74.1 & 63.8 & 73.4 & 67.1 & 67.0 & 48.0 & 71.4 & 6.7 & 4.6 & 5.4 \\
 & BF-SI-TI-DIM (ours) & \textbf{89.3} & \textbf{87.3} & \textbf{84.1} & \textbf{83.9} & \textbf{83.3} & \textbf{82.4} & \textbf{78.3} & \textbf{86.4} & \textbf{9.7} & \textbf{7.7} & \textbf{9.2} \\
 \hline
\multirow{8}{*}{Incres-v2} & TI-DIM & 48.0 & 43.8 & 38.7 & 44.6 & 41.1 & 57.4 & 24.9 & 42.8 & 3.4 & 1.6 & 1.9 \\
& PI-TI-DI-FGSM & 49.9 & 51.2 & 46.0 & 40.9 & 45.3 & 78.1 & 41.4 & 47.6 & 6.5 & 5.4 & 5.6 \\
 & SI-TI-DIM & 79.0 & 76.1 & 73.6 & 75.7 & 73.3 & 68.2 & 52.5 & 75.6 & 6.7 & 5.4 & 6.0 \\
 & VT-TI-DIM & 65.7 & 61.1 & 59.0 & 60.4 & 57.8 & 61.3 & 37.2 & 60.6 & 4.7 & 2.8 & 3.4 \\
 & FI-TI-DIM & 58.3 & 54.4 & 53.9 & 52.5 & 53.1 & 57.0 & 40.1 & 57.3 & 6.6 & 3.9 & 4.3 \\
& BF-TI-DIM (ours) & \textbf{83.1} & \textbf{82.3} & \textbf{82.3} & \textbf{80.3} & \textbf{80.9} & \textbf{79.8} & \textbf{74.3} & \textbf{82.6} & \textbf{10.5} & \textbf{8.1} & \textbf{9.4} \\
\cline{2-13}
& Admix-TI-DIM & 85.3 & 82.0 & 79.5 & 82.4 & 79.6 & 74.2 & 59.7 & 82.4 & 7.9 & 6.0 & 5.7 \\
 & BF-SI-TI-DIM (ours) & \textbf{91.4} & \textbf{90.3} & \textbf{89.9} & \textbf{88.2} & \textbf{88.8} & \textbf{85.3} & \textbf{84.4} & \textbf{89.4} & \textbf{11.8} & \textbf{9.6} & \textbf{12.4} \\
 \hline
\multirow{8}{*}{Res-152} & TI-DIM & 55.0 & 53.6 & 43.1 & 55.7 & 46.7 & 61.2 & 32.4 & 52.3 & 4.6 & 3.4 & 3.4 \\
& PI-TI-DI-FGSM & 54.4 & 56.9 & 45.6 & 43.8 & 46.1 & 78.2 & 47.8 & 49.2 & 7.9 & 5.9 & 6.5 \\
 & SI-TI-DIM & 77.3 & 76.5 & 67.0 & 73.3 & 68.4 & 70.9 & 53.2 & 72.6 & 6.8 & 5.5 & 5.2 \\
 & VT-TI-DIM & 64.5 & 61.3 & 55.0 & 60.7 & 54.8 & 68.2 & 41.3 & 59.9 & 6.1 & 5.0 & 4.9 \\
 & FI-TI-DIM & 70.2 & 66.0 & 59.4 & 64.0 & 61.1 & 71.4 & 47.7 & 66.0 & 8.4 & 6.2 & 5.5 \\
 & BF-TI-DIM (ours) & \textbf{87.6} & \textbf{86.4} & \textbf{81.1} & \textbf{83.9} & \textbf{81.8} & \textbf{84.0} & \textbf{77.7} & \textbf{85.0} & \textbf{12.0} & \textbf{10.4} & \textbf{9.9} \\
  \cline{2-13}
  & Admix-TI-DIM & 83.7 & 81.4 & 73.7 & 81.2 & 77.0 & 75.0 & 59.5 & 80.1 & 8.3 & 6.2 & 6.3 \\
 & BF-SI-TI-DIM (ours) & \textbf{90.9} & \textbf{90.5} & \textbf{85.9} & \textbf{85.9} & \textbf{86.3} & \textbf{87.1} & \textbf{84.8} & \textbf{89.4} & \textbf{12.3} & \textbf{10.8} & \textbf{10.1} \\
 \hline
\end{tabular}
}
\label{tab:single}
\end{table*}

\subsection{Boundary Analysis}
\label{sec:3.4}

In previous Section~\ref{sec:3.2}, we have demonstrated that gradients of different models on decision boundary are closer. Based on this observation, we craft adversarial examples by boundary gradient with the aim of narrowing the gap between the substitute model and victim model. Considering that the perturbations are added to the original input, it is interesting to see whether this way is effective in moving adversarial examples out of the boundary of the victim model. 
Therefore, we compared the $\mathcal{L}_\infty$-distance to the decision boundary (decision boundary distance) along the direction of the sign gradient generated by different approaches to further support the rationality of our method design.

The results are shown in Table~\ref{tab:l2explantation}. Remarkably, our BF-FGSM can effectively reduce the decision boundary distance of the victim model compared to state-of-the-art attacks. For example, in Inc-v3$\rightarrow$IncRes-v2 case, 
the average decision boundary distance along the direction of sign gradient derived from our BF-FGSM is only 19.23, while existing input transformation methods (\textit{e.g.}, Admix) usually have much longer distances.These results again demonstrate that taking boundary gradients into account is helpful for boosting adversarial attacks. 

\begin{table}[H]
\centering
\caption{The average $\mathcal{L}_\infty$-decision boundary distance of transformers and CNNs along natural directions and adversarial directions respectively.}
\resizebox{0.8\linewidth}{!}{
\begin{tabular}{ccc}
\hline
Models & Natural Directions & Adversarial Directions \\
\hline
\hline
Inc-v3 & 66.36 & 26.25 \\
Inc-v4 & 68.46 & 33.77 \\
ViT-B & 83.30 & 26.43 \\
Deit-B & 81.65 & 23.84 \\
\hline
\end{tabular}
}
\label{tab:transcnnboundary}
\end{table}

In addition, we observe that model robustness is positively correlated with the decision boundary distance.
Therefore, we conjecture that the decision boundary distance obtained along arbitrary directions is larger than that of CNNs, which enables transformers are more robust than CNNs. 
To verify this conjecture, we compare the average decision boundary distance of transformers and CNNs along natural directions (sampled by Gauss distribution) and adversarial directions (the gradient of the source image).
The results are shown in Tab.~\ref{tab:transcnnboundary}, the average decision boundary distance of transformers is larger than that of CNNs when along natural directions, while it is similar when along the adversarial directions (\textit{i.e.} one-iteration white-box attack). 
A larger decision boundary distance means a smaller adversarial subspace that intersects with other models, thus adversarial examples are harder to transfer to transformers compared to CNNs, so transformers exhibit stronger robustness than CNNs.
Therefore we claim that \textit{model robustness is positively correlated with the average decision boundary distance along natural directions.}

\section{Experiment}

In this section, we display the experimental results to demonstrate the effectiveness of our proposed method. In Sec.\ref{sec:4.1}, we first define the experimental setup. Then we conduct experiments to verify that the proposed method is effective for both normally trained models and defense models in Sec.\ref{sec:4.2} and Sec.\ref{sec:4.3}. 
In Sec.\ref{sec:4.4}, we further conduct experiments on transformers and compare them with CNNs.
Finally, we conduct a series of ablation experiments to study the impact of different parameters in Sec.~\ref{sec:4.5}.

\begin{table*}[h]
\centering
\caption{The attack success rates (\%) of black-box attacks against eleven defenses. The adversarial examples are crafted via an ensemble of Inc-v3, Inc-v4, IncRes-v2 and Res-152 and the weight for each model is 1/4. Note that Admix is equipped with SI-FGSM by default.}
\resizebox{1\linewidth}{!}{
\begin{tabular}{c|c|M{1cm}|M{1cm}|M{1cm}|M{1cm}|M{1cm}|M{1cm}|M{1cm}|M{1cm}|M{1cm}|M{1cm}}
\hline
  Attacks & Inc-v3$_{ens3}$ & Inc-v3$_{ens4}$ & IncRes-v2$_{ens}$ & HGD & R\&P & RS & NRP & NIPS-r3 & Res152$_{B}$ & Res152$_{D}$ & ResNeXt$_{DA}$\\
\hline
\hline
 TI-DIM & 79.6 & 77.7 & 68.2 & 80.2 & 73.8 & 68.6 & 43.5 & 76.5 & 5.6 & 4.4 & 4.8 \\
 PI-TI-DIM & 75.0 & 76.4 & 68.2 & 69.2 & 67.9 & 84.2 & 61.5 & 72.8 & 10.0 & 8.0 & 8.9 \\
 SI-TI-DIM & 93.3 & 92.5 & 88.9 & 92.4 & 91.4 & 83.4 & 75.9 & 92.2 & 9.4 & 7.7 & 9.6 \\
 VT-TI-DIM & 87.4 & 85.7 & 82.7 & 86.4 & 82.8 & 75.0 & 60.1 & 84.3 & 7.8 & 6.2 & 6.7 \\
 FI-TI-DIM & 83.4 & 83.6 & 75.2 & 85.1 & 77.2 & 73.5 & 61.3 & 79.3 & 7.6 & 6.3 & 6.4 \\
 Admix-TI-DIM & 93.9 & 92.9 & 90.3 & 94.0 & 91.3 & 82.4 & 76.0 & 92.0 & 10.5 & 8.3 & 9.9 \\
 BF-TI-DIM (Ours) & \textbf{95.0} & \textbf{93.9} & \textbf{93.0} & \textbf{94.1} & \textbf{92.5} & \textbf{90.8} & \textbf{89.1} & \textbf{94.1} & \textbf{15.6} & \textbf{13.9} & \textbf{14.9}\\
\hline
\end{tabular}}
\label{tab:ensemble}
\end{table*}

\subsection{Experiment Setup}
\label{sec:4.1}
\noindent {\bfseries{Dataset.}} Following prior works~\cite{mifgsm,difgsm,admix}, we use the ImageNet-compatible dataset\footnote{\url{https://github.com/tensorflow/cleverhans/tree/master/examples/nips17_adversarial_competition/dataset}} comprised of 1000 images to conduct experiments.

\noindent {\bfseries{Models.}} For evaluation of transferability, we adopt six normally trained models: Inception-v3 (Inc-v3)~\cite{inc-v3}, Inception-v4 (Inc-v4), Inception-Resnet-v2 (IncRes-v2)~\cite{inc-v4}, Resnet-v2-50 (Res-50), Resnet-v2-101 (Res-101), Resnet-v2-152 (Res-152)~\cite{res152} and eleven defense models: Inc-v3$_{ens3}$, Inc-v3$_{ens4}$, IncRes-v2$_{ens}$~\cite{ens}, HGD~\cite{HGD}, R\&P~\cite{RP}, RS~\cite{RS}, NRP~\cite{NRP}, NIPS-r3\footnote{\url{https://github.com/anlthms/nips-2017/blob/master/poster/defense.pdf}}, ResNeXt$_{DA}$, Res152$_{B}$, Res152$_{D}$~\cite{FeatureDeoise}. 
For evaluating transformers, we adopt four models: Vision Transformer (ViT)~\cite{vit}, Data-efficient Image Transformers (DeiT)~\cite{deit}, Transformer in Transformer (TNT)~\cite{tnt} and Swin Transformer (Swin)~\cite{swin}.

\noindent {\bfseries{Competitor.}} To manifest the effectiveness of our proposed approach, we compare it with state-of-the-art attacks including MI-FGSM~\cite{mifgsm}, DI-FGSM~\cite{difgsm}, PI-FGSM~\cite{pifgsm}, SI-FGSM~\cite{sinifgsm}, VT-FGSM~\cite{vtfgsm}, FI-FGSM~\cite{fia} and Admix~\cite{admix}. We also compare the combined version of these attacks, \textit{e.g.}, DIM (\textit{i.e.}, combined version of MI-FGSM and DI-FGSM and TI-DIM.

\noindent {\bfseries{Parameter Settings.}} In all experiments, the maximum perturbation $\epsilon = 16$, the iteration $T = 10$, and the step size $\alpha = \epsilon/T = 1.6$. For MI-FGSM, we set the decay factor $\mu$ = 1.0. For DI-FGSM, we set the transformation probability $p$ = 0.5. For TI-FGSM, we set the kernel length $k$ = 7. For PI-FGSM, we set the amplification factor $\beta$ = 10, project factor $\gamma$ = 16 and the kernel length $k_w$ = 3 for normally trained models, $k_w$ = 7 for defense models. For SI-FGSM, we set the number of copies $m$ = 5. For VT-FGSM, we set the hyper-parameter $\beta$ = 1.5, number of sampling examples $N=20$. For FI-FGSM, the drop probability $p_d$ = 0.3 for normally trained models and $p_d$ = 0.1 for defense models, and the ensemble number $N=30$. For Admix, we set sample number $m_2$ = 3 and the admix ratio $\eta$ = 0.2.  For our proposed BF-FGSM, we set maximum number of moves $t=5$ to guarantee efficiency, the shrinkage factor  $\gamma$ = 0.6, the initial direction standard deviation $\sigma=20$, and the number of sampled boundary points $N=20$. Note that the parameter settings for the combined version are the same.

\begin{table*}[h]
\centering
\caption{CNN to transformers. The attack success rates (\%) of black-box attacks against ten normally trained transformers. The adversarial examples are crafted via Inc-v3, Inc-v4, IncRes-v2 and Res-152 respectively. Note that Admix is equipped with SI-FGSM by default.}
\resizebox{1\linewidth}{!}{
\begin{tabular}{c|c|c|c|c|c|c|c|c|c|c|c|c}
\hline
Models & Attacks & ViT-S & ViT-B & ViT-L & DeitT-T & DeiT-S & DeiT-B & TNT-S & Swin-T & Swin-S & Swin-B & Average \\
\hline
\hline
\multirow{8}{*}{Inc-v3} & MI-FGSM & 20.8 & 17.5 & 11.6 & 40.7 & 32.8 & 24.2 & 25.5 & 42.1 & 35.6 & 24.9 & 27.6 \\
 & DI-MI-FGSM & 29.3 & 24.1 & 16.2 & 46.5 & 39.8 & 33.3 & 36.5 & 51.0 & 44.4 & 32.5 & 35.4 \\
 & TI-MI-FGSM & 25.2 & 20.0 & 14.6 & 48.3 & 41.2 & 28.4 & 27.9 & 38.6 & 41.1 & 28.6 & 31.4 \\
 & SI-MI-FGSM & 33.4 & 26.0 & 16.2 & 52.8 & 40.8 & 33.9 & 38.0 & 51.1 & 44.3 & 32.4 & 36.9 \\
 & VMI & 33.5 & 28.0 & 17.7 & 50.0 & 43.2 & 38.4 & 39.5 & 54.9 & 49.2 & 39.0 & 39.3 \\
 & BF-MI-FGSM & \textbf{55.3} & \textbf{44.2} & \textbf{29.4} & \textbf{68.8} & \textbf{58.6} & \textbf{55.8} & \textbf{60.5} & \textbf{71.4} & \textbf{64.7} & \textbf{48.2} & \textbf{55.7} \\
 \cline{2-13}
 & Admix & 38.0 & 29.7 & 19.1 & 58.3 & 46.9 & 40.7 & 44.5 & 60.2 & 54.5 & 38.5 & 43.0 \\
 & BF-SIM & \textbf{63.8} & \textbf{46.8} & \textbf{35.2} & \textbf{75.5} & \textbf{65.4} & \textbf{59.4} & \textbf{66.0} & \textbf{71.8} & \textbf{68.0} & \textbf{53.4} & \textbf{60.5} \\
 \hline
\multirow{8}{*}{Inc-v4} & MI-FGSM & 21.6 & 17.3 & 12.4 & 40.2 & 34.4 & 28.0 & 24.6 & 43.6 & 35.3 & 26.7 & 28.4 \\
 & DI-MI-FGSM & 27.4 & 23.0 & 16.5 & 26.2 & 39.4 & 32.9 & 36.8 & 53.9 & 46.9 & 32.1 & 33.5 \\
 & TI-MI-FGSM & 25.2 & 21.0 & 15.8 & 49.5 & 43.0 & 28.9 & 29.2 & 41.8 & 45.9 & 32.8 & 33.3 \\
 & SI-MI-FGSM & 37.8 & 32.0 & 21.5 & 55.3 & 47.6 & 45.5 & 46.0 & 59.7 & 53.5 & 42.2 & 44.1 \\
 & VMI & 35.1 & 28.1 & 18.8 & 51.4 & 47.0 & 40.6 & 44.7 & 58.3 & 51.0 & 39.3 & 41.4 \\
 & BF-MI-FGSM & \textbf{57.0} & \textbf{47.3} & \textbf{35.0} & \textbf{71.0} & \textbf{61.9} & \textbf{59.3} & \textbf{64.0} & \textbf{72.4} & \textbf{68.1} & \textbf{52.0} & \textbf{58.8} \\
 \cline{2-13}
 & Admix & 43.8 & 35.5 & 23.3 & 60.9 & 53.9 & 47.8 & 52.0 & 67.2 & 60.7 & 48.0 & 49.3 \\
 & BF-SIM & \textbf{70.7} & \textbf{56.8} & \textbf{43.0} & \textbf{81.2} & \textbf{72.4} & \textbf{67.8} & \textbf{72.1} & \textbf{78.8} & \textbf{76.3} & \textbf{59.9} & \textbf{67.9} \\
 \hline
\multirow{8}{*}{IncRes-v2} & MI-FGSM & 21.8 & 18.1 & 12.8 & 41.8 & 32.1 & 25.2 & 27.2 & 48.2 & 47.1 & 27.6 & 30.2 \\
 & DI-MI-FGSM & 29.8 & 24.5 & 15.8 & 47.5 & 39.1 & 35.4 & 36.9 & 58.4 & 53.1 & 35.2 & 37.6 \\
 & TI-MI-FGSM & 28.8 & 23.2 & 17.7 & 51.0 & 44.3 & 32.4 & 29.6 & 47.0 & 50.0 & 31.0 & 35.5 \\
 & SI-MI-FGSM & 44.1 & 35.7 & 25.1 & 58.9 & 50.9 & 45.0 & 47.9 & 64.6 & 61.4 & 44.3 & 47.8 \\
 & VMI & 37.5 & 29.2 & 22.1 & 49.9 & 44.2 & 40.8 & 42.9 & 60.1 & 57.8 & 43.7 & 42.8 \\
 & BF-MI-FGSM & \textbf{60.5} & \textbf{49.4} & \textbf{38.2} & \textbf{68.5} & \textbf{61.6} & \textbf{60.1} & \textbf{63.8} & \textbf{70.4} & \textbf{67.6} & \textbf{51.5} & \textbf{59.2} \\
 \cline{2-13}
 & Admix & 51.7 & 40.7 & 26.5 & 65.6 & 57.9 & 51.7 & 54.1 & 72.1 & 68.4 & 52.3 & 54.1 \\
 & BF-SIM & \textbf{73.0} & \textbf{60.4} & \textbf{47.6} & \textbf{78.0} & \textbf{72.3} & \textbf{70.3} & \textbf{71.8} & \textbf{75.5} & \textbf{74.1} & \textbf{60.1} & \textbf{68.3} \\
 \hline
\multirow{8}{*}{Res-152} & MI-FGSM & 21.4 & 17.8 & 11.8 & 38.3 & 25.5 & 23.1 & 23.7 & 40.6 & 35.4 & 31.7 & 26.9 \\
 & DI-MI-FGSM & 37.2 & 26.4 & 15.5 & 50.8 & 39.0 & 37.5 & 39.5 & 60.6 & 52.7 & 41.8 & 40.1 \\
 & TI-MI-FGSM & 33.8 & 27.2 & 18.9 & 53.3 & 42.9 & 36.1 & 34.5 & 48.5 & 46.3 & 33.2 & 37.5 \\
 & SI-MI-FGSM & 25.6 & 18.3 & 13.3 & 44.1 & 28.9 & 28.1 & 27.5 & 45.5 & 40.5 & 32.7 & 30.5 \\
 & VMI & 38.9 & 30.9 & 21.0 & 54.8 & 41.0 & 39.8 & 42.9 & 60.8 & 56.9 & 45.0 & 43.2 \\
 & BF-MI-FGSM & \textbf{61.5} & \textbf{45.2} & \textbf{33.5} & \textbf{74.1} & \textbf{59.9} & \textbf{60.1} & \textbf{61.5} & \textbf{74.5} & \textbf{66.5} & \textbf{53.7} & \textbf{59.1} \\
 \cline{2-13}
 & Admix & 28.1 & 21.4 & 14.9 & 46.9 & 31.1 & 30.0 & 32.4 & 52.8 & 45.9 & 35.5 & 33.9 \\
 & BF-MI-SIM & \textbf{59.3} & \textbf{44.0} & \textbf{32.9} & \textbf{74.2} & \textbf{59.3} & \textbf{56.6} & \textbf{57.2} & \textbf{70.2} & \textbf{64.6} & \textbf{50.6} & \textbf{56.9}\\
 \hline
\end{tabular}
}

\label{tab:cnntotrans}
\end{table*}

\subsection{Attack Normally trained CNN Models}
\label{sec:4.2}

In this section, we investigate the vulnerability of normally trained models. We first compare MI-FGSM, DI-FGSM, PI-FGSM with our BF-FGSM to verify the effectiveness of our method, and the results are shown in Table~\ref{tab:normal}. Notably, our proposed BF-FGSM consistently surpasses all well-known methods in the black-box setting. For example, when attacking Inc-v4, BF-FGSM can outperform MI-FGSM, DI-FGSM and PI-FGSM by 7.9\%, 17.1\% and 12.6\% on average, respectively.

Furthermore, we compare three other state-of-the-art attacks that are equipped with momentum term~\cite{mifgsm}, \textit{i.e.}, SI-MI-FGSM, VT-MI-FGSM and FI-MI-FGSM. From the Table~\ref{tab:normal}, we observe that momentum term can significantly boost our method. Specifically, it raises the average success rate from 68.3\% to 89.0\%. Remarkably, our BF-MI-FGSM can obtain an average success rate of 92.8\% when the substitute model is Res-152, which outperforms SI-MI-FGSM, VT-MI-FGSM and FI-MI-FGSM by 10.0\%, 10.2\% and 3.9\%. These results also convincingly demonstrate the superiority of our approach.

\begin{table*}[h]
\centering
\caption{transformers to transformers. The attack success rates (\%) of black-box attacks against ten normally trained transformers. The adversarial examples are crafted via ViT-B, DeiT-B, TNT-S and Swin-B respectively. Note that Admix is equipped with SI-FGSM by default. ``*" indicates white-box attacks.}
\resizebox{1\linewidth}{!}{
\begin{tabular}{c|c|c|c|c|c|c|c|c|c|c|c|c}
\hline
Models & Attack & ViT-S & ViT-B & ViT-L & DeitT-T & DeiT-S & DeiT-B & TNT-S & Swin-T & Swin-S & Swin-B & Average \\
\hline
\hline
\multirow{8}{*}{ViT-B} & MI-FGSM & 91.2 & \textbf{100.0*} & 85.9 & 70.3 & 75.6 & 80.7 & 60.2 & 54.1 & 56.0 & 58.4 & 73.2 \\
 & DI-MI-FGSM & 75.9 & 94.8* & 62.4 & 59.1 & 58.4 & 63.4 & 58.7 & 53.6 & 53.9 & 55.2 & 63.5 \\
  & TI-MI-FGSM & 87.1 & 99.9* & 82.8 & 66.2 & 69.4 & 73.2 & 54.6 & 45.8 & 45.8 & 46.0 & 67.1 \\
 & SI-MI-FGSM & 95.4 & 99.9* & 91.7 & 85.8 & 88.4 & 89.6 & 76.3 & 67.1 & 69.3 & 68.7 & 83.2 \\
 & VMI & 95.8 & \textbf{100.0*} & 94.2 & 79.6 & 84.2 & 88.2 & 77.6 & 69.0 & 71.3 & 75.9 & 83.6 \\
 & BF-MI-FGSM & \textbf{98.4} & 99.6* & \textbf{97.8} & \textbf{91.7} & \textbf{95.4} & \textbf{97.0} & \textbf{93.5} & \textbf{84.4} & \textbf{87.5} & \textbf{88.2} & \textbf{93.4} \\
 \cline{2-13}
 & Admix & 92.6 & 99.9* & 87.7 & 74.0 & 80.7 & 83.6 & 67.8 & 60.9 & 64.3 & 64.0 & 77.6 \\
 & BF-MI-SIM & \textbf{97.8} & 99.0* & \textbf{96.6} & \textbf{95.0} & \textbf{96.5} & \textbf{96.5} & \textbf{91.8} & \textbf{83.5} & \textbf{83.2} & \textbf{84.1} & \textbf{92.4} \\
 \hline
\multirow{8}{*}{DeiT-B} & MI-FGSM & 83.0 & \textbf{76.1} & 57.8 & 87.9 & 94.9 & \textbf{100.0*} & 80.8 & 73.0 & 69.8 & 53.2 & 77.7 \\
 & DI-MI-FGSM & 57.2 & 50.3 & 39.5 & 68.1 & 70.1 & 90.3* & 68.1 & 61.9 & 58.0 & 47.5 & 61.1 \\
 & TI-MI-FGSM & 77.2 & 66.5 & 54.6 & 86.2 & 92.0 & \textbf{100.0} & 76.9 & 59.6 & 57.7 & 39.2 & 71.0 \\
 & SI-MI-FGSM & 93.0 & 86.0 & 76.7 & \textbf{95.6} & \textbf{98.1} & \textbf{100.0*} & 92.2 & 83.0 & 81.3 & 67.4 & 87.3 \\
 & VMI & 89.5 & 86.8 & 74.5 & 91.9 & 97.6 & \textbf{100.0*} & 90.5 & 83.4 & 81.9 & 68.8 & 86.5 \\
 & BF-MI-FGSM & \textbf{94.4} & \textbf{91.6} & \textbf{87.6} & 95.0 & 97.6 & 99.7* & \textbf{93.0} & \textbf{88.6} & \textbf{88.4} & \textbf{81.3} & \textbf{91.7} \\
 \cline{2-13}
 & Admix & 83.4 & 75.8 & 60.0 & 88.9 & 94.6 & 99.8* & 84.2 & 77.6 & 75.7 & 57.5 & 79.8 \\
 & BF-MI-SIM & \textbf{96.5} & \textbf{93.6} & \textbf{89.6} & \textbf{98.2} & \textbf{98.4} & \textbf{100.0*} & \textbf{95.1} & \textbf{91.2} & \textbf{89.6} & \textbf{82.1} & \textbf{93.4} \\
 \hline
\multirow{8}{*}{TNT-S} & MI-FGSM & 59.7 & 45.3 & 30.9 & 75.0 & 78.7 & 74.1 & \textbf{100.0*} & 68.9 & 62.8 & 46.4 & 64.2 \\
 & DI-MI-FGSM & 60.1 & 49.7 & 36.4 & 69.2 & 70.2 & 69.2 & 85.0* & 70.4 & 68.3 & 51.8 & 63.0 \\
  & TI-MI-FGSM & 68.2 & 54.1 & 39.7 & 80.2 & 81.2 & 75.8 & \textbf{100.0} & 66.3 & 60.5 & 42.3 & 66.8 \\
 & SI-MI-FGSM & 75.1 & 59.8 & 42.5 & 85.8 & 90.3 & 86.0 & \textbf{100.0*} & 80.6 & 73.4 & 58.4 & 75.2 \\
 & VMI & 83.7 & 71.3 & 56.0 & 89.6 & 91.6 & 89.3 & \textbf{100.0*} & 86.5 & 83.6 & 73.4 & 82.5 \\
  & BF-MI-FGSM & \textbf{90.4} & \textbf{82.2} & \textbf{72.7} & \textbf{92.9} & \textbf{94.6} & \textbf{93.4} & 99.6* & \textbf{91.8} & \textbf{90.6} & \textbf{83.2} & \textbf{89.1} \\
  \cline{2-13}
 & Admix & 65.1 & 49.8 & 32.7 & 76.3 & 81.0 & 77.0 & \textbf{100.0*} & 75.4 & 69.7 & 52.4 & 67.9 \\
 & BF-MI-SIM & \textbf{93.2} & \textbf{86.1} & \textbf{76.7} & \textbf{96.2} & \textbf{96.4} & \textbf{95.5} & 99.7* & \textbf{93.3} & \textbf{92.8} & \textbf{85.6} & \textbf{91.6} \\
 \hline
\multirow{8}{*}{Swin-B} & MI-FGSM & 33.4 & 30.3 & 21.3 & 40.7 & 34.7 & 35.5 & 34.5 & 80.5 & 84.5 & \textbf{100.0*} & 49.5 \\
 & DI-MI-FGSM & 29.5 & 28.8 & 20.0 & 35.2 & 28.5 & 30.7 & 29.2 & 38.6 & 41.2 & 59.7* & 34.1 \\
 & TI-MI-FGSM & 39.8 & 34.0 & 25.9 & 50.8 & 50.9 & 45.8 & 42.3 & 63.4 & 70.4 & \textbf{100.0} & 52.3 \\
 & SI-MI-FGSM & 48.5 & 43.0 & 28.8 & 51.4 & 48.0 & 49.3 & 48.2 & 88.4 & 90.7 & \textbf{100.0*} & 59.6 \\
 & VMI & 53.2 & 51.4 & 39.3 & 52.6 & 51.8 & 57.8 & 54.3 & 83.6 & 87.1 & \textbf{100.0*} & 63.1 \\
  & BF-MI-FGSM & \textbf{78.0} & \textbf{73.9} & \textbf{70.0} & \textbf{75.2} & \textbf{77.6} & \textbf{81.1} & \textbf{76.9} & \textbf{87.5} & \textbf{89.8} & 96.4* & \textbf{80.6}\\
  \cline{2-13}
 & Admix & 32.5 & 28.6 & 18.8 & 39.5 & 31.6 & 35.6 & 35.1 & 76.2 & 81.7 & \textbf{100.0*} & 48.0 \\
& BF-MI-SIM & \textbf{84.7} & \textbf{77.8} & \textbf{71.6} & \textbf{83.0} & \textbf{85.5} & \textbf{86.7} & \textbf{83.7} & \textbf{95.3} & \textbf{96.0} & 99.3* & \textbf{86.4}\\
  \cline{2-13}
 \hline
\end{tabular}
}
\label{tab:trans2trans}
\end{table*}

\subsection{Attack Defense CNN Models}
\label{sec:4.3}

Although many adversarial attack methods can successfully fool normally trained models, they usually fail at attacking models with defense mechanisms. 

To further validate the superiority of our method, we conduct a series of experiments against defense models. Specifically, we compare our BF-TI-DIM and BF-SI-TI-DIM with TI-DIM, PI-TI-DIM, SI-TI-DIM, FI-TI-DIM and Admix-TI-DIM. Note that Admix is equipped with SI-FGSM by default.

{\bfseries{Single-Model Attacks.}} We first craft adversarial examples via a single model, and the results are shown in Table~\ref{tab:single}. Due to the space limitation, here we only report the results of attacking Inc-v3 and other results can be found in Supplementary Sec. A. 
From Table~\ref{tab:single}, we observe that the transferability of adversarial examples crafted by our methods is far superior to existing state-of-the-art approaches. In particular, when transferring adversarial examples to NRP, SI-TI-DIM obtains the highest attack success rate (40.1\%) among other methods, while our BF-TI-DIM can achieve a success rate of 69.9\%.

In addition, our BF-TI-DIM can be further enhanced when combined with the SI-FGSM. Compared with Admix-TI-DIM, whose average success rate is 47.9\%, our BF-SI-TI-DIM can significantly enhance the transferability up to 62.8\%. 

These results also demonstrate that exploring the update direction at the decision boundary can help the adversarial example find an effective path to bypass the defense mechanisms.

{\bfseries{Ensemble-Model Attacks.}} To further improve the transferability, we craft adversarial examples via an ensemble of models. Specifically, we adopt the strategy proposed in~\cite{mifgsm} which fuses the logit activations of different models. 
As demonstrated in Table~\ref{tab:ensemble}, our adversarial examples crafted via an ensemble of Inc-v3, Inc-v4, IncRes-v2 and Res-152 can evade most defense models with a high success rate. Noticeably, 95.0\% adversarial examples crafted by our BF-TI-DIM can fool Inc-v3$_{ens3}$. On average, our proposed attack can fool state-of-the-art defense models at a 71.5\% success rate, which is 18.5\%, 16.8\%, 4.6\%, 11.1\%, 4.1\% and 13.4\% higher than TI-DIM, PI-TI-DIM, SI-TI-DIM, VT-TI-DIM, Admix-TI-DIM and FI-TI-DIM, respectively. 
Admittedly, feature denoising defenses~\cite{FeatureDeoise} are very robust to current attacks. Nonetheless, our method still outperforms state-of-the-art PI-TI-DIM and Admix-TI-DIM by 5.8\% and 5.2\%. These experimental results also show that most current defenses are still vulnerable to malicious adversarial examples.

\begin{table*}[h]
\centering
\caption{transformers to CNNs. The attack success rates (\%) of black-box attacks against six normally trained CNNs. The adversarial examples are crafted via ViT-B, DeiT-B, TNT-S and Swin-B respectively. Note that Admix is equipped with SI-FGSM by default.}
\resizebox{0.9\linewidth}{!}{
\begin{tabular}{c|c|M{1cm}|M{1cm}|M{1cm}|M{1cm}|M{1cm}|M{1cm}|M{1cm}}
\hline
Models & Attacks & Inc-v3 & Inc-v4 & IncRes-v2 & Res-50 & Res-101 & Res-152 & Average \\
\hline
\hline
\multirow{8}{*}{ViT-B} & MI-FGSM & 45.4 & 41.6 & 36.9 & 71.1 & 64.8 & 59.5 & 53.2 \\
 & DI-MI-FGSM & 43.5 & 38.1 & 32.3 & 61.2 & 54.5 & 51.5 & 46.9 \\
 & TI-MI-FGSM & 48.7 & 44.0 & 35.8 & 69.4 & 64.0 & 59.8 & 53.6 \\
 & SI-MI-FGSM & 60.2 & 55.5 & 47.8 & 78.9 & 75.8 & 70.9 & 64.9 \\
 & VMI & 57.0 & 54.6 & 48.6 & 76.7 & 71.1 & 69.1 & 62.9 \\
 & BF-MI-FGSM (ours) & \textbf{79.5} & \textbf{75.7} & \textbf{72.0} & \textbf{85.3} & \textbf{85.5} & \textbf{84.2} & \textbf{80.4} \\
 \cline{2-9}
 & Admix & 52.0 & 46.9 & 39.3 & 72.9 & 68.2 & 63.9 & 57.2 \\
 & BF-SIM (ours) & \textbf{81.5} & \textbf{78.7} & \textbf{76.3} & \textbf{88.6} & \textbf{88.4} & \textbf{87.4} & \textbf{83.5} \\
 \hline
\multirow{8}{*}{DeiT-B} & MI-FGSM & 56.6 & 50.5 & 43.0 & 76.7 & 73.2 & 68.2 & 61.4 \\
 & DI-MI-FGSM & 46.7 & 40.1 & 32.8 & 61.6 & 56.4 & 53.8 & 48.6 \\
 & TI-MI-FGSM & 57.9 & 52.0 & 43.4 & 76.9 & 71.8 & 66.1 & 61.4 \\
 & SI-MI-FGSM & 68.7 & 61.5 & 54.4 & 84.2 & 82.4 & 76.0 & 71.2 \\
 & VMI & 64.2 & 59.6 & 54.0 & 81.4 & 78.8 & 73.7 & 68.6 \\
 & BF-MI-FGSM (ours) & \textbf{76.3} & \textbf{73.7} & \textbf{69.7} & \textbf{87.4} & \textbf{85.6} & \textbf{84.0} & \textbf{79.5} \\
 \cline{2-9}
 & Admix & 59.9 & 52.1 & 46.5 & 78.5 & 73.5 & 69.9 & 63.4 \\
 & BF-SIM (ours) & \textbf{81.9} & \textbf{77.5} & \textbf{76.1} & \textbf{90.9} & \textbf{88.9} & \textbf{86.6} & \textbf{83.7} \\
 \hline
\multirow{8}{*}{TNT-S} & MI-FGSM & 52.0 & 44.6 & 39.0 & 76.3 & 69.3 & 65.6 & 57.8 \\
 & DI-MI-FGSM & 53.0 & 47.8 & 43.5 & 67.5 & 63.3 & 60.0 & 55.9 \\
 & TI-MI-FGSM & 58.4 & 50.3 & 45.9 & 78.3 & 70.9 & 66.7 & 61.8 \\
 & SI-MI-FGSM & 62.1 & 55.1 & 51.7 & 81.8 & 76.3 & 74.0 & 66.8 \\
 & VMI & 68.2 & 65.2 & 61.2 & 83.1 & 80.2 & 78.0 & 72.7 \\
 & BF-MI-FGSM (ours) & \textbf{80.8} & \textbf{78.4} & \textbf{76.1} & \textbf{86.6} & \textbf{85.8} & \textbf{84.8} & \textbf{82.1} \\
 \cline{2-9}
 & Admix & 55.6 & 49.3 & 43.5 & 77.8 & 72.6 & 68.9 & 61.3 \\
 & BF-SIM (ours) & \textbf{84.7} & \textbf{82.5} & \textbf{80.8} & \textbf{90.6} & \textbf{89.0} & \textbf{87.9} & \textbf{85.9} \\
 \hline
\multirow{8}{*}{Swin-B} & MI-FGSM & 32.2 & 26.7 & 19.1 & 54.5 & 49.2 & 47.1 & 38.1 \\
 & DI-MI-FGSM & 26.2 & 22.8 & 15.5 & 44.6 & 41.9 & 37.2 & 31.4 \\
 & TI-MI-FGSM & 43.7 & 41.2 & 30.0 & 68.0 & 58.3 & 55.2 & 49.4 \\
 & SI-MI-FGSM & 41.1 & 34.2 & 28.1 & 60.1 & 55.9 & 55.1 & 45.8 \\
 & VMI & 43.1 & 41.0 & 33.8 & 60.6 & 55.0 & 54.8 & 48.1 \\
 & BF-MI-FGSM (ours) & \textbf{66.9} & \textbf{66.5} & \textbf{62.2} & \textbf{77.2} & \textbf{75.4} & \textbf{74.2} & \textbf{70.4} \\
 \cline{2-9}
 & Admix & 30.1 & 24.5 & 19.7 & 53.3 & 48.3 & 47.1 & 37.2 \\
 & BF-SIM (ours) & \textbf{75.1} & \textbf{71.8} & \textbf{68.0} & \textbf{84.6} & \textbf{82.8} & \textbf{82.0} & \textbf{77.4} \\
 \hline
\end{tabular}
}
\label{tab:tran2cnn}
\end{table*}

\subsection{Attack Transformers}
\label{sec:4.4}

Most of the current adversarial attack methods only attack the CNN model, however, transformer is now known to have higher accuracy and robustness. Therefore, we compare our algorithm with recently state-of-the-art attacks on transformers to disclose the vulnerability of transformers and show the effectiveness of our algorithm. The experiments have three parts: CNNs to transformers, transformers to transformers and transformers to CNNs. For CNNs, we choose Inc-v3, Inc-v4, IncRes-v2, Res-50, Res-101 and Res-152. For transformers, we choose several versions of ViT, DeiT, TNT and Swin (\textit{i.e.} ViT-S, ViT-B, ViT-L, DeiT-T, DeiT-S, DeiT-B, TNT-S, Swin-T, Swin-S, Swin-B).

{\bfseries{CNNs to Transformers.}} We first craft adversarial examples via CNN models and leverage the adversarial examples to attack transformers.  The results are shown in Table~\ref{tab:cnntotrans}. Compare with Table~\ref{tab:normal}, when using our BF-MI-FGSM, the average success rate from Inc-v3 to transformers is 32.7\% lower than that of Inc-v3 to CNNs.
The results reveal that transformers are more robust than CNNs, as demonstrated in \cite{vitrobust}. However, our method still has a considerate success rate and outperforms other attacks by a large margin. Specifically, our BF-MI-FGSM fool transformers at a 58.8\% success rate on average when attacking Inc-v4, which is 30.4\%, 25.3\%, 25.5\%, 14.7\%, 17.4\% and 9.5\% higher than MI-FGSM, DI-MI-FGSM, TI-MI-FGSM, SI-MI-FGSM, VMI and Admix respectively. Combined with SIM, our BF-SIM has an even higher success rate of 67.9\% when attacking Inc-v4.

{\bfseries{Transformers to Transformers.}} To demonstrate our method, we also craft adversarial examples via transformers and leverage the adversarial examples to attack transformers. 
The results are shown in Table~\ref{tab:trans2trans}. 
Compare to Table~\ref{tab:cnntotrans}, the attack success rate improves by a large margin, which shows that transformers can generate more transferable adversarial examples.
In this case, our method gets higher attack success rate compare with other state-of-the-art methods.
Our BF-MI-FGSM obtains an 88.7\% average attack success rate which outperforms MI-FGSM, DI-MI-FGSM, TI-MI-FGSM, SI-MI-FGSM, VMI and Admix by 22.6\%, 33.2\%, 24.4\%, 12.3\%, 9.8\% and 20.4\% respectively.

\begin{figure}[t]
    \centering
    \includegraphics[width=8.25cm]{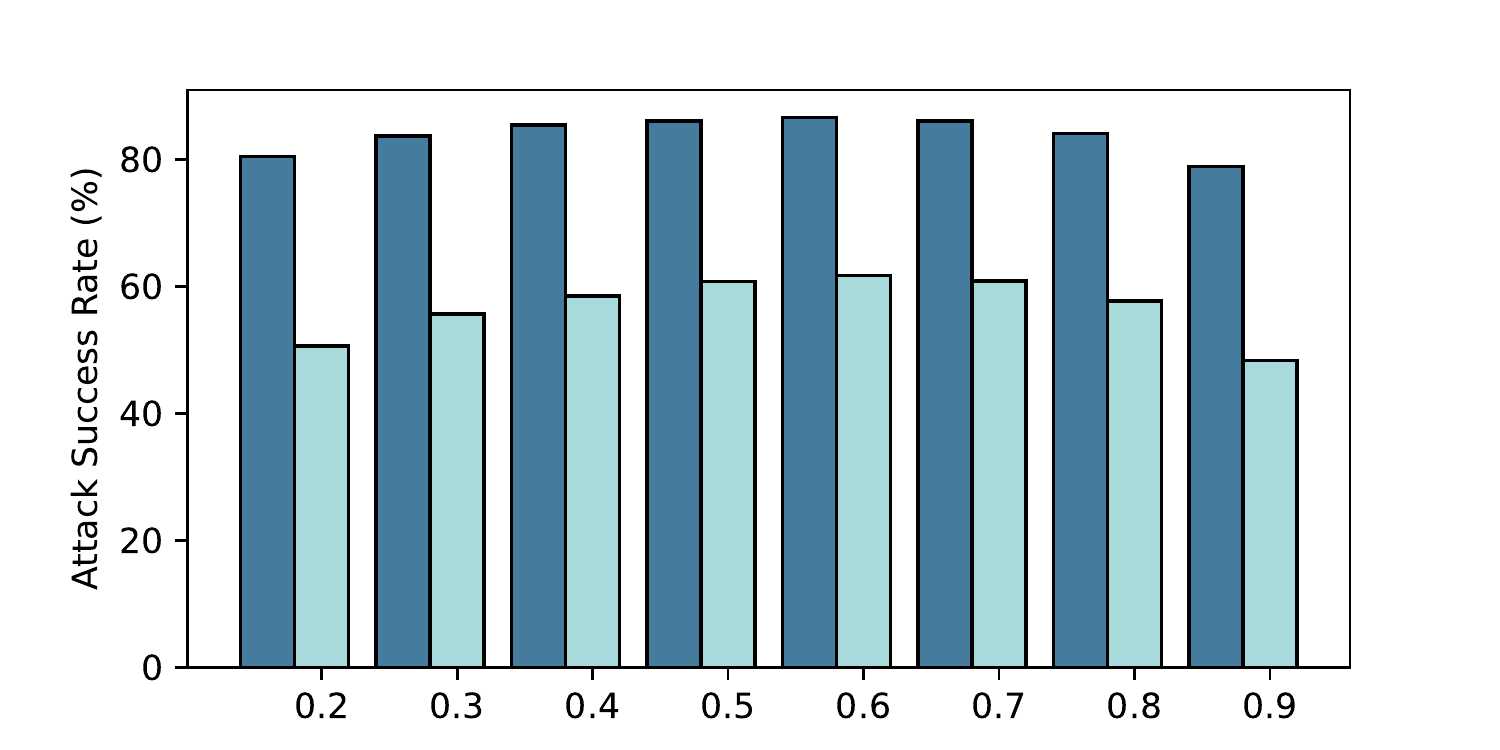}
    \caption{Average attack success rate (\%) of adversarial examples crafted by BF-MI-FGSM w.r.t. shrinkage factor $\gamma$. The substitute model is Inc-v3. \textbf{Left}: The transferability towards normally trained models. \textbf{Right}: The transferability towards defense models.}
    \label{fig:Gamma}
\end{figure}

\begin{figure}[t]
    \centering
    \includegraphics[width=8.25cm]{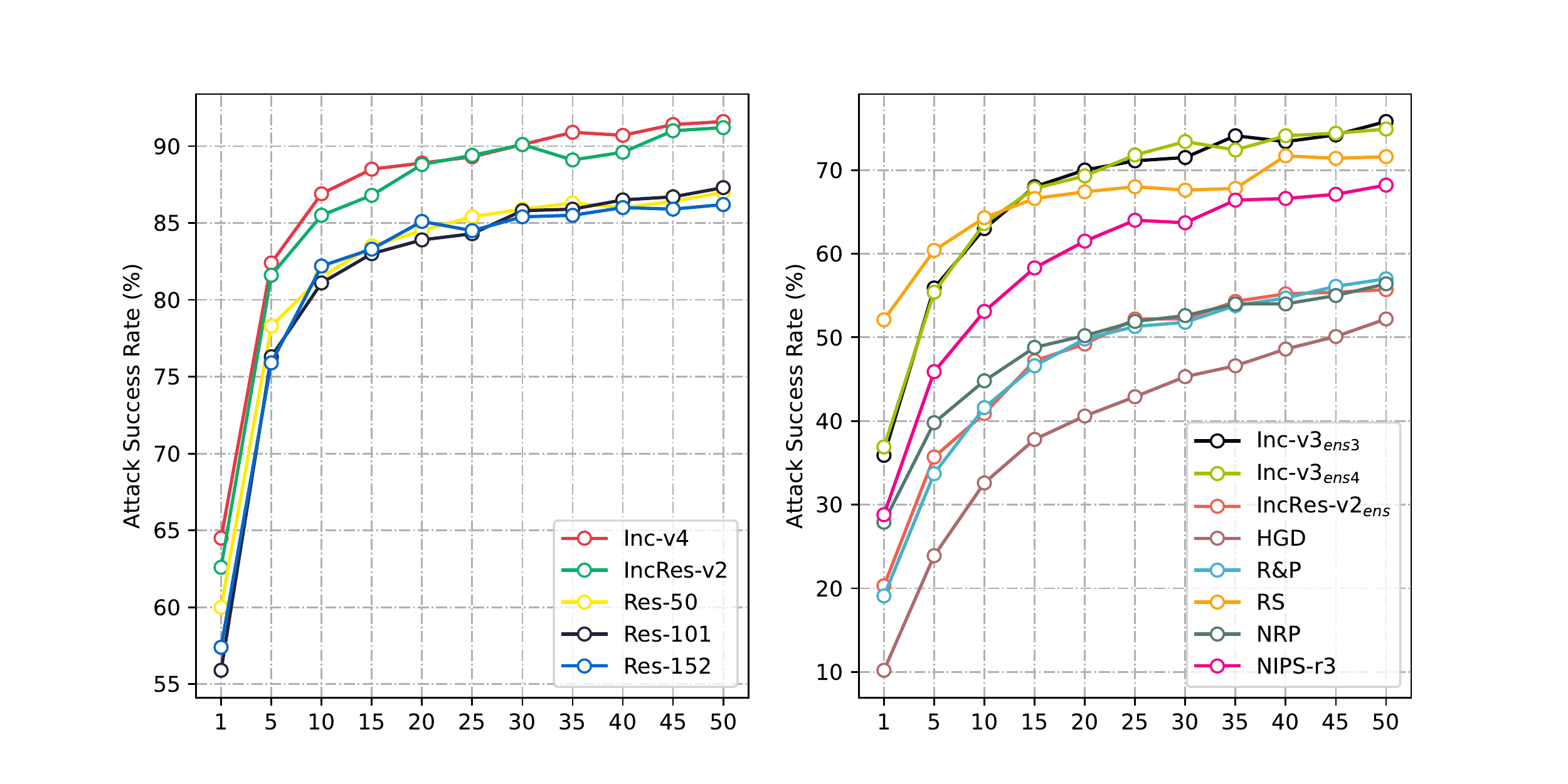}
    \caption{The attack success rates (\%) of adversarial examples crafted by BF-MI-FGSM w.r.t. sampled number $N$. The substitute model is Inc-v3.  \textbf{Left}: The transferability towards normally trained models. \textbf{Right}: The transferability towards defense models.}
    \label{fig:N}
\end{figure}

{\bfseries{Transformers to CNNs.}} In this section, we craft adversarial examples via transformers and leverage the adversarial examples to attack CNNs. 
The results are shown in Table~\ref{tab:tran2cnn}, which shows that adversarial examples crafted by our method are more transferable than other method.
Notably, our BF-MI-FGSM obtains a 78.1\% average attack success rate, which outperforms other state-of-the-art method by 13.6\%.
When combined with SIM, our BF-SIM get a 82.6\% average attack success rate.

\begin{figure*}[h]
    \centering
    \includegraphics[width = 1.0\linewidth]{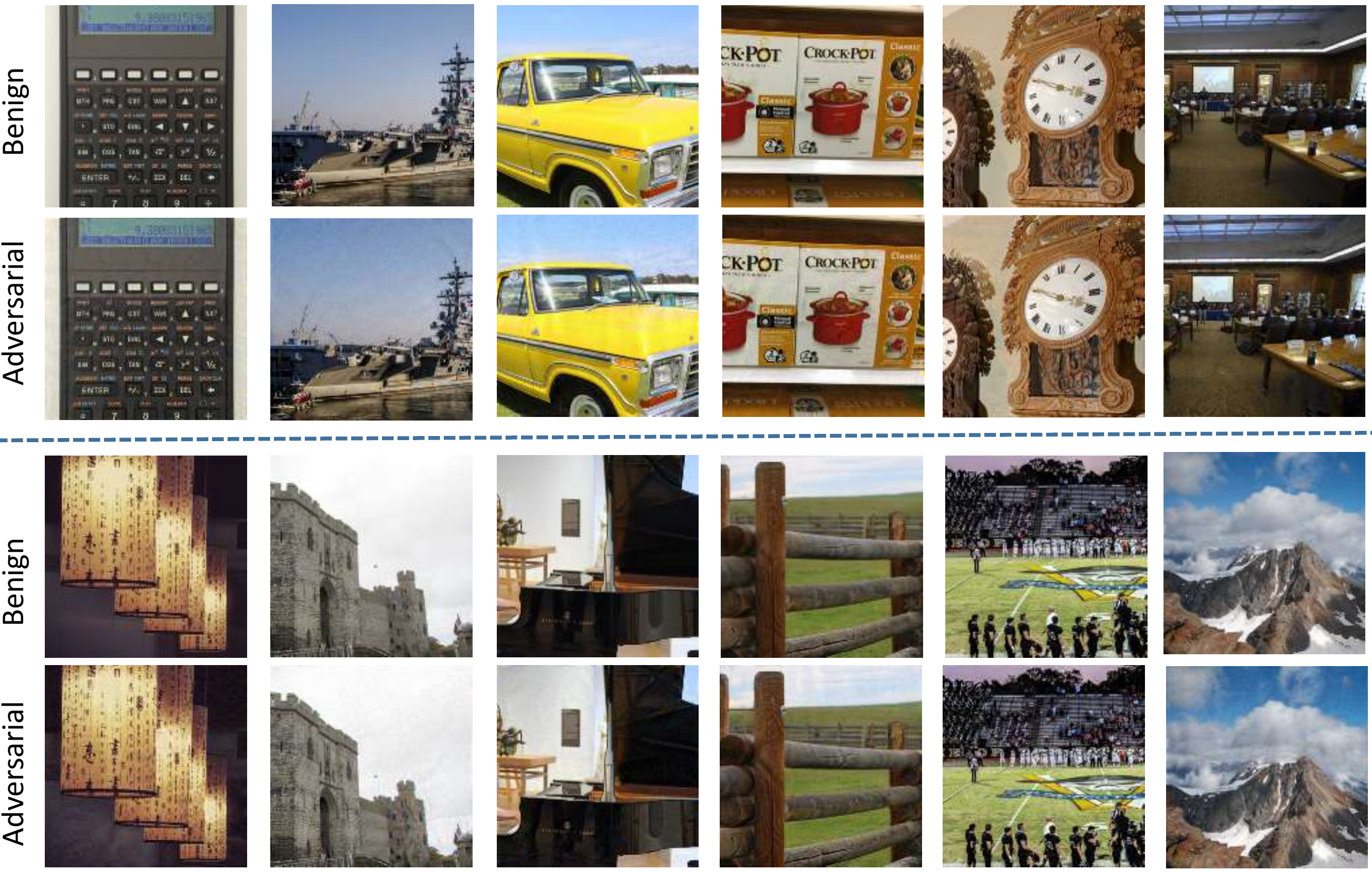}
    \caption{Visualization of randomly selected benign images and corresponding adversarial examples.}
    \label{pic1}
\end{figure*}

\begin{figure*}[h]
    \centering
    \includegraphics[width=18cm]{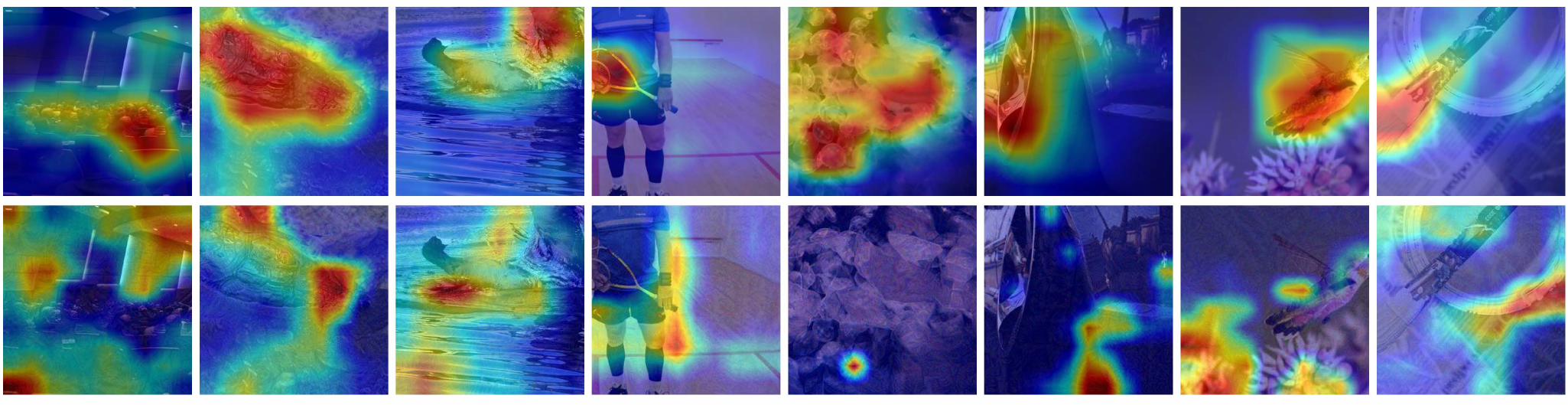}
    \caption{Visualization for attention shift. We use Grad-CAM~\cite{grad_cam} to visualize attention maps of clean (1st row) and adversarial images (2nd row). Adversarial examples are crafted via Inc-v3 by our BF-FGSM. The comparison demonstrates that our method is capable of shifting model's attention on images.}
    \label{fig:attention}
\end{figure*}

\subsection{Ablation Study}
\label{sec:4.5}

In this section, we conduct ablation experiments to analyze the impact of our parameters on transferability. 

{\bfseries{Effect of shrinkage factor $\gamma$.}} To guarantee the efficiency of generating adversarial examples, we fix the maximum move steps $t$ to 5. In this case, a small value of $\gamma$ may pull obtained data points too close to the inputs, and conversely, a large value of $\gamma$ may cause data points to be far away from the decision boundary.
Therefore, it is necessary to select a suitable shrinkage factor $\gamma$ for boundary search. As shown in Figure~\ref{fig:Gamma}, the average attack success rate rises as $\gamma$ increases and then drops after $\gamma$ exceeds 0.6. Therefore, we set the shrinkage factor $\gamma$ to 0.6 in all experiments.

{\bfseries{Effect of sampled number $N$.}} In this section, we investigate the effect of the number of sampled boundary points for our BF-MI-FGSM, and adversarial examples are crafted via Inc-v3 with $\gamma=0.6$. As illustrated in Figure~\ref{fig:N}, when transferring to the normally trained model, 
the increment in transferability is very limited after $N$ exceeds 20. In contrast, for the defense model, the transferability of the adversarial examples continues to be enhanced even after $N$ exceeds 20. This suggests that conducting extensive exploration on the decision boundary of a normally trained model can simulate defense models to some extent. However, the computational cost is proportional to $N$. Therefore, we set $N$ = 20 in our paper for a trade-off.

\section{Visualization}

\subsection{Visualization of Adversarial Examples}
We randomly select twelve benign images and their corresponding adversarial examples crafted by our BF-FGSM in Figure~\ref{pic1}. It can be observed that these
crafted adversarial examples are human-imperceptible.

\subsection{Visualization of Attention shift}
In this section, we investigate the effectiveness of our attack from a perspective of attention shift on adversarial examples. As illustrated in Figure~\ref{fig:attention}, our proposed method effectively narrows the original attention region and enhances the irrelevant region. Consequently, the victim model will capture other irrelevant features, thus leading to misclassification.

\section{Conclusion}

This paper gives a new insight into the boundary gradients of different models and shows that they are more similar than the original gradients. Furthermore, as the decision boundary tends to be flat, we conjecture that the input image reaches the decision boundary closer along the direction of the boundary gradients.
Based on it, we proposed a Boundary Fitting Attack, which enhances the transferability of adversarial examples by leveraging the decision boundary information of the substitute model. 
We also introduce a concept: decision boundary distance, and propose that model robustness is positively related to the decision boundary distance along natural directions.
Extensive experiments demonstrate the effectiveness of our method and support the rationality of our method design.
Moreover, we also find that transformers are more robust than CNNs and our method is able to craft more transferable adversarial examples on transformers than existing methods.
Finally, our method achieves the highest average attack success rate on both CNNs and transformers.

\bibliographystyle{IEEEtran}
\bibliography{tip}

\end{document}


\maketitle

\appendix

\section{Attack defense models by single model}
In section 4.4, we conduct single-model attacks against defense models. However, due to space limitation we only include the results of Inc-v3 as the substitute model. Here we show the rest of the experimental results in Table~\ref{tab:singlerest}. 

From Table~\ref{tab:singlerest} we can observe that the proposed BF-TI-DIM and BF-SI-TI-DIM exceed all the state-of-the-art attack methods. Notably, our BAT-TI-DIM has an attack success rate at 58.9\% on average, which is 33.3\%, 27.2\%, 13.0\%, 18.0\%, 8.1\% and 18.2\% higher than TI-DIM, PI-TI-DIM, SI-TI-DIM, VT-TI-DIM, Admix-TI-DIM and FI-TI-DIM respectively. Furthermore, when combined with SI-FGSM, the proposed BAT-SI-TI-DIM obtains better results. BAT-SI-TI-DIM gets a significant 67.4\% success rate on average when the adversarial examples crafted via IncRes-v2.

\begin{table*}[h]
\resizebox{1\linewidth}{!}{
\begin{tabular}{c|c|c|c|c|c|c|c|c|c|c|c|c}
\hline
 Models & Attacks & Inc-v3_{ens3} & Inc-v3_{ens4} & IncRes-v2_{ens} & HGD & R\&P & RS & NRP & NIPS-r3 & Res152_{B} & Res152_{D} & ResNeXt_{DA}\\
\hline
\hline
\multirow{8}{*}{Inc-v4} & TI-DIM & 38.4 & 38.5 & 27.6 & 34.1 & 29.4 & 55.3 & 19.1 & 33.2 & 2.9 & 1.7 & 1.6 \\
 & PI-TI-DI-FGSM & 42.7 & 44.2 & 32.8 & 33.4 & 34.0 & 75.0 & 32.9 & 37.2 & 5.8 & 4.2 & 6.4 \\
 & SI-TI-DIM & 70.3 & 67.5 & 57.3 & 64.6 & 58.2 & 67.8 & 41.4 & 62.3 & 5.3 & 4.6 & 4.8 \\
 & VT-TI-DIM & 57.9 & 57.5 & 46.7 & 56.1 & 58.0 & 59.4 & 41.6 & 62.4 & 3.5 & 3.0 & 3.3 \\
 & FI-TI-DIM & 61.4 & 58.0 & 51.2 & 54.2 & 51.8 & 62.7 & 39.0 & 55.9 & 6.0 & 3.8 & 3.8 \\
 & BF-TI-DIM & \textbf{83.7} & \textbf{82.1} & \textbf{74.9} & \textbf{77.3} & \textbf{76.2} & \textbf{79.2} & \textbf{70.9} & \textbf{78.6} & \textbf{9.1} & \textbf{7.4} & \textbf{8.4} \\
 \cline{2-13}
  & Admix-TI-DIM & 77.3 & 74.1 & 63.8 & 73.4 & 67.1 & 67.0 & 48.0 & 71.4 & 6.7 & 4.6 & 5.4 \\
 & BF-SI-TI-DIM & \textbf{89.3} & \textbf{87.3} & \textbf{84.1} & \textbf{83.9} & \textbf{83.3} & \textbf{82.4} & \textbf{78.3} & \textbf{86.4} & \textbf{9.7} & \textbf{7.7} & \textbf{9.2} \\
 \hline
\multirow{8}{*}{Incres-v2} & TI-DIM & 48.0 & 43.8 & 38.7 & 44.6 & 41.1 & 57.4 & 24.9 & 42.8 & 3.4 & 1.6 & 1.9 \\
& PI-TI-DI-FGSM & 49.9 & 51.2 & 46.0 & 40.9 & 45.3 & 78.1 & 41.4 & 47.6 & 6.5 & 5.4 & 5.6 \\
 & SI-TI-DIM & 79.0 & 76.1 & 73.6 & 75.7 & 73.3 & 68.2 & 52.5 & 75.6 & 6.7 & 5.4 & 6.0 \\
 & VT-TI-DIM & 65.7 & 61.1 & 59.0 & 60.4 & 57.8 & 61.3 & 37.2 & 60.6 & 4.7 & 2.8 & 3.4 \\
 & FI-TI-DIM & 58.3 & 54.4 & 53.9 & 52.5 & 53.1 & 57.0 & 40.1 & 57.3 & 6.6 & 3.9 & 4.3 \\
& BF-TI-DIM & \textbf{83.1} & \textbf{82.3} & \textbf{82.3} & \textbf{80.3} & \textbf{80.9} & \textbf{79.8} & \textbf{74.3} & \textbf{82.6} & \textbf{10.5} & \textbf{8.1} & \textbf{9.4} \\
\cline{2-13}
& Admix-TI-DIM & 85.3 & 82.0 & 79.5 & 82.4 & 79.6 & 74.2 & 59.7 & 82.4 & 7.9 & 6.0 & 5.7 \\
 & BF-SI-TI-DIM & \textbf{91.4} & \textbf{90.3} & \textbf{89.9} & \textbf{88.2} & \textbf{88.8} & \textbf{85.3} & \textbf{84.4} & \textbf{89.4} & \textbf{11.8} & \textbf{9.6} & \textbf{12.4} \\
 \hline
\multirow{8}{*}{Res-152} & TI-DIM & 55.0 & 53.6 & 43.1 & 55.7 & 46.7 & 61.2 & 32.4 & 52.3 & 4.6 & 3.4 & 3.4 \\
& PI-TI-DI-FGSM & 54.4 & 56.9 & 45.6 & 43.8 & 46.1 & 78.2 & 47.8 & 49.2 & 7.9 & 5.9 & 6.5 \\
 & SI-TI-DIM & 77.3 & 76.5 & 67.0 & 73.3 & 68.4 & 70.9 & 53.2 & 72.6 & 6.8 & 5.5 & 5.2 \\
 & VT-TI-DIM & 64.5 & 61.3 & 55.0 & 60.7 & 54.8 & 68.2 & 41.3 & 59.9 & 6.1 & 5.0 & 4.9 \\
 & FI-TI-DIM & 70.2 & 66.0 & 59.4 & 64.0 & 61.1 & 71.4 & 47.7 & 66.0 & 8.4 & 6.2 & 5.5 \\
 & BF-TI-DIM & \textbf{87.6} & \textbf{86.4} & \textbf{81.1} & \textbf{83.9} & \textbf{81.8} & \textbf{84.0} & \textbf{77.7} & \textbf{85.0} & \textbf{12.0} & \textbf{10.4} & \textbf{9.9} \\
  \cline{2-13}
  & Admix-TI-DIM & 83.7 & 81.4 & 73.7 & 81.2 & 77.0 & 75.0 & 59.5 & 80.1 & 8.3 & 6.2 & 6.3 \\
 & BF-SI-TI-DIM & \textbf{90.9} & \textbf{90.5} & \textbf{85.9} & \textbf{85.9} & \textbf{86.3} & \textbf{87.1} & \textbf{84.8} & \textbf{89.4} & \textbf{12.3} & \textbf{10.8} & \textbf{10.1} \\
 \hline
\end{tabular}}
\caption{The attack success rates (\%) of black-box attacks against eleven defenses. The adversarial examples are crafted via Inc-v4, Incres-v2 and Res-152 respectively. Note that Admix is equipped with SI-FGSM by default.}
\label{tab:singlerest}
\end{table*}

\section{Visualization}

\subsection{Visualization of Adversarial Examples}
\begin{figure*}[ht]
    \centering
    \includegraphics[width = 1.0\linewidth]{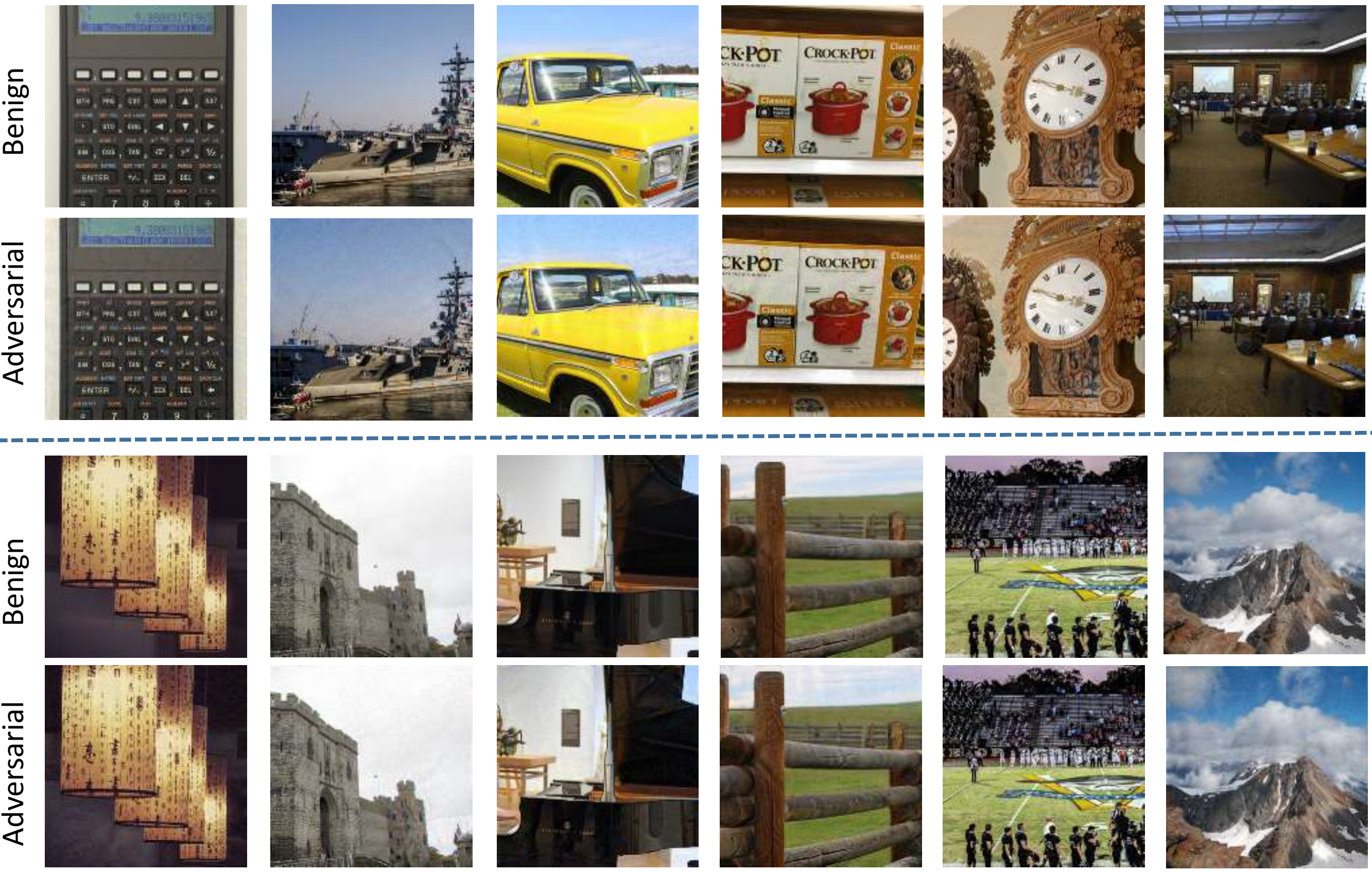}
    \caption{Visualization of randomly selected benign images and corresponding adversarial examples.}
    \label{pic1}
\end{figure*}

We randomly select twelve benign images and their corresponding adversarial examples crafted by our BF-FGSM in Figure~\ref{pic1}. It can be observed that these
crafted adversarial examples are human-imperceptible.

\subsection{Visualization of Attention shift}
\begin{figure*}[h]
    \centering
    \includegraphics[width=18cm]{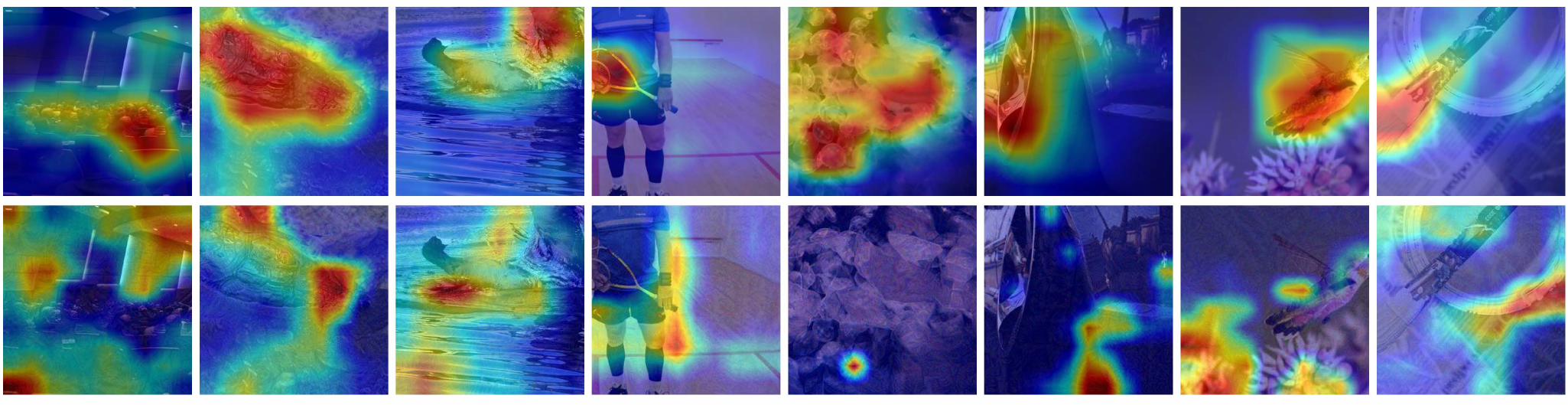}
    \caption{Visualization for attention shift. We use Grad-CAM~\cite{grad_cam} to visualize attention maps of clean (1st row) and adversarial images (2nd row). Adversarial examples are crafted via Inc-v3 by our BF-FGSM. The comparison demonstrates that our method is capable of shifting model's attention on images.}
    \label{fig:attention}
\end{figure*}

In this section, we investigate the effectiveness of our attack from a perspective of attention shift on adversarial examples. As illustrated in Figure~\ref{fig:attention}, our proposed method effectively narrows the original attention region and enhances the irrelevant region. Consequently, the victim model will capture other irrelevant features, thus leading to misclassification.

\clearpage

\bibliographystyle{named}
\bibliography{ijcai22}